\documentclass[lettersize,journal]{IEEEtran}
\usepackage{amsmath,amsfonts}
\usepackage{algorithmic}
\usepackage{algorithm}
\usepackage{array}
\usepackage[caption=false,font=normalsize,labelfont=sf,textfont=sf]{subfig}
\usepackage{textcomp}
\usepackage{stfloats}
\usepackage{url}
\usepackage{verbatim}
\usepackage{graphicx}
\usepackage{multirow}
\usepackage{amsthm}
\usepackage{booktabs}
\usepackage{color}
\usepackage{subfloat}
\usepackage{hyperref}
\hypersetup{hidelinks}
\usepackage[switch]{lineno}
\usepackage{cite}

\begin{document}

\title{Illumination Distillation Framework for Nighttime Person Re-Identification and A New Benchmark}
\author{Andong Lu,
		Zhang Zhang,~\IEEEmembership{Member, IEEE},
		Yan Huang,
		Yifan Zhang,
		Chenglong Li,\\
		Jin Tang,
      and Liang Wang,~\IEEEmembership{Fellow, IEEE}

    \thanks{
    This work was supported in part by the Natural Science Foundation of Anhui Province (No. 2208085J18),
    in part by the National Natural Science Foundation of China (No. 62106006),
    in part by the Natural Science Foundation of Anhui Higher Education Institution (No. 2022AH040014),
in part by the National Key R\&D Program of China (No. 2022ZD0117901),
in part by the Peak Discipline Construction Project (Computer Science and Technology) (No. Z010111016),
in part by the Fellowship of China Postdoctoral Science Foundation (No. 2022T150698),
and in part by the University Synergy Innovation Program of Anhui Province, China (No. GXXT-2022-033).

    A. Lu and J. Tang are with Anhui Provincial Key Laboratory of Multimodal Cognitive Computation, School of Computer Science and Technology, Anhui University, Hefei 230601, China. A. Lu is also with the National Laboratory of Pattern Recognition, Institute of Automation, Chinese Academy of Sciences, Beijing 100190, China.
    (e-mail: adlu\_ah@foxmail.com; tangjin@ahu.edu.cn)
    
    C. Li is with Information Materials and Intelligent Sensing Laboratory of Anhui Province, Anhui Provincial Key Laboratory of Multimodal Cognitive Computation, School of Artificial Intelligence, Anhui University, Hefei 230601, China. 
    (e-mail: lcl1314@foxmail.com)
    
    Z. Zhang, Y. Huang, Y. Zhang and L. Wang are with the National Laboratory of Pattern Recognition, Institute of Automation, Chinese Academy of Sciences, Beijing 100190, China.
    (e-mail: zzhang@nlpr.ia.ac.cn; huangyan.750@outlook.com, yifanzhang.cs@gmail.com; wangliang@nlpr.ia.ac.cn)
    }      
    }

\markboth{Journal of \LaTeX\ Class Files,~Vol.~14, No.~8, August~2021}%
{Shell \MakeLowercase{\textit{et al.}}: A Sample Article Using IEEEtran.cls for IEEE Journals}

\maketitle

\begin{abstract}
Nighttime person Re-ID (person re-identification in the nighttime) is a very important and challenging task for visual surveillance but it has not been thoroughly investigated. Under the low illumination condition, the performance of person Re-ID methods usually  sharply deteriorates. To address the low illumination challenge in nighttime person Re-ID, this paper proposes an Illumination Distillation Framework (IDF), which utilizes illumination enhancement and illumination distillation schemes to promote the learning of Re-ID models. Specifically, IDF consists of a master branch, an illumination enhancement branch, and an illumination distillation module. The master branch is used to extract the features from a nighttime image. The illumination enhancement branch first estimates an enhanced image from the nighttime image using a nonlinear curve mapping method and then extracts the enhanced features. However, nighttime and enhanced features usually contain data noise due to unstable lighting conditions and enhancement failures. To fully exploit the complementary benefits of nighttime and enhanced features while suppressing data noise, we propose an illumination distillation module. In particular, the illumination distillation module fuses the features from two branches through a bottleneck fusion model and then uses the fused features to guide the learning of both branches in a distillation manner. In addition, we build a real-world nighttime person Re-ID dataset, named \emph{Night600}, which contains 600 identities captured from different viewpoints and nighttime illumination conditions under complex outdoor environments. Experimental results demonstrate that our IDF can achieve state-of-the-art performance on two nighttime person Re-ID datasets (\emph{i.e.}, \emph{Night600} and \emph{Knight} ). We will release our code and dataset at \href{https://github.com/Alexadlu/IDF}{https://github.com/Alexadlu/IDF}.
\end{abstract}


\section{Introduction}
\begin{figure}[t]
    \centering
    \begin{tabular}[b]{cc}
    \includegraphics[scale=0.40]{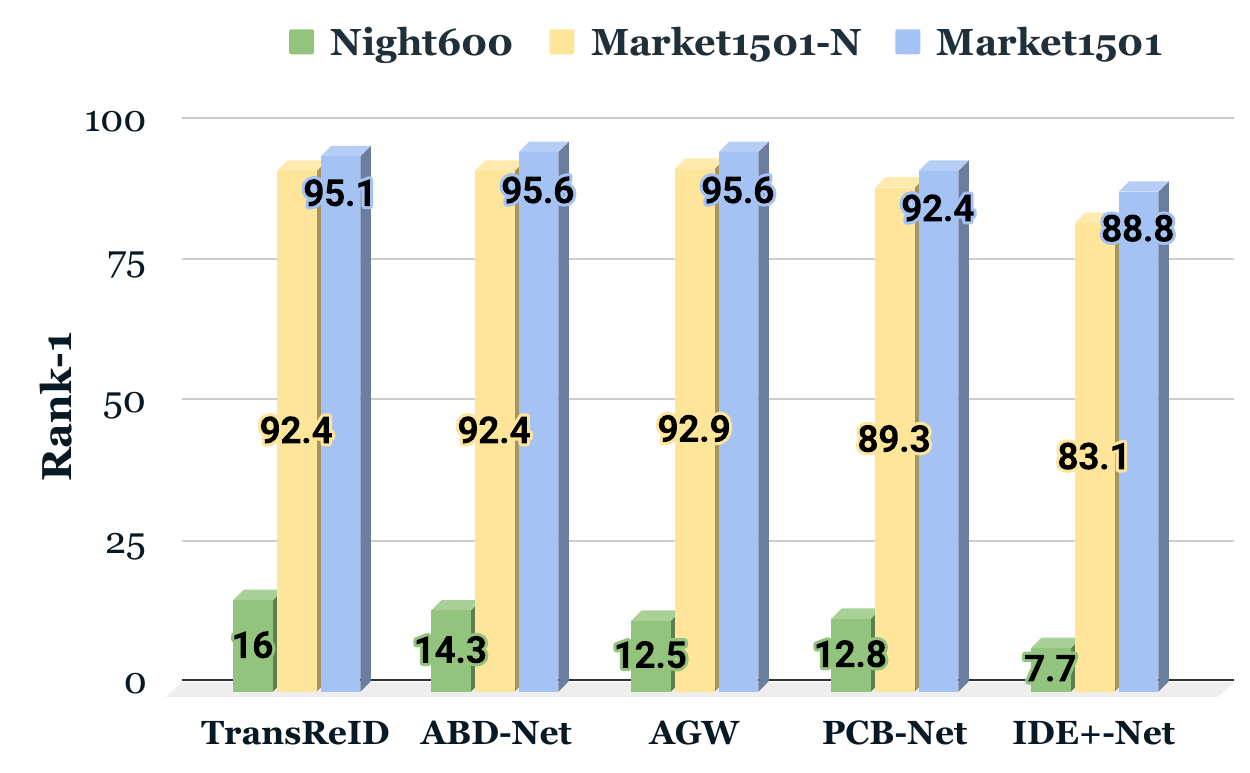}
       \vspace{-.5cm}
    \end{tabular} 
    \caption{Evaluation results of several advanced Re-ID algorithms including TransReID~\cite{TransReID_2021}, ABD-Net~\cite{ABDNet_2019}, AGW~\cite{AGW_2021}, PCB-Net\cite{PCB_2018} and IDE+-Net\cite{IDE+_2019} on the proposed nighttime \emph{Night600} dataset, synthetic low-illumination \emph{Market1501-N} dataset and original daytime \emph{Market1501} dataset.}
    \label{fig:baselines}
\end{figure}

\IEEEPARstart{T}{he} darkness of night is a natural camouflage of criminal suspects and a recent report~\cite{night_report} reveals that violent crimes are more likely to occur at night.
Therefore, nighttime visual analysis has critical application potential in public security and forensics.
Person re-identification (Re-ID) aims at identifying a person of interest across multiple non-overlap camera views, which is necessary for the study of nighttime visual analysis,~\emph{e.g.}, analysing person traces.
%
%
However, existing studies~\cite{gao2020dcr,CPA_tifs,AIIS_pami,CR_tip,adaptive_ill_2019TMM,BoT_2019,jia2022learning,zhang2020person,ABDNet_2019,AGW_2021,TransReID_2021} mainly focus on person Re-ID in daytime scenarios, and the performance of daytime person Re-ID has thus been greatly boosted.
But the imaging quality degrades significantly at night, which significantly hinders the application of existing algorithms.
As shown in Fig.~\ref{fig:baselines}, existing advanced person Re-ID methods perform poorly on the nighttime person Re-ID (person Re-ID in the nighttime) dataset, \emph{Night600}.
Therefore, nighttime person Re-ID is a more important but challenging task.

The quality of person images collected at night is usually low because of low illumination.
Some recent works~\cite{Real_world_CVPR2020,Ill_Invariant_2019ACMMM,IllAP_2020icip, adaptive_ill_2019TMM} investigate the issue of illumination change person Re-ID by learning illumination invariant features.
%
For instance, Zeng~\emph{et al.}~\cite{adaptive_ill_2019TMM} and Huang~\emph{et al.}~\cite{Real_world_CVPR2020} use generative adversarial networks (GAN) to learn disentangled representations, which are invariant to illumination changes.
These works use gamma correlation to synthesize low-illumination images from daytime ones, but it is hard to simulate the low illumination in the nighttime.
As shown in Fig.~\ref{fig:baselines}, compared with the performance on the daytime dataset Market1501~\cite{market1501_2015}, the models have little degradation on the synthetic low-illumination dataset (Market1501-N). 
This suggests that synthetic low-illumination data are not able to reflect the real challenges posed by nighttime data. %
In contrast, the performance of these models on our nighttime Re-ID dataset dramatically degrades.

To deal with this problem, a straightforward approach is to improve the visibility of nighttime images through existing illumination enhancement methods. Benefiting from the powerful feature representation of deep neural networks, illumination enhancement techniques~\cite{zhang2019kindling,lu2020tbefn,yang2020fidelity,triantafyllidou2020low,wang2020lightening, DCE_2020,dce++tpami} achieve tremendous development. 
Existing methods according to different learning strategies can be divided into two categories: supervised learning-based and unsupervised learning-based. Among them, supervised learning-based illumination enhancement  works~\cite{zhang2019kindling,lu2020tbefn,yang2020fidelity} are the current mainstream, but they rely on large-scale paired training data for learning. However, it is extremely difficult to obtain paired data in real-world scenarios. Therefore, some studies~\cite{triantafyllidou2020low,wang2020lightening} use synthetic techniques to build paired data for training, but they do not yield satisfactory results due to the domain gap between real and synthetic data. In order to avoid paired data limitations, some researchers propose unsupervised illumination enhancement methods~\cite{DCE_2020,dce++tpami}, they usually design various reference-free losses to guide model optimization. 
However, these methods focus on improving the quality of visual perception, thus using them as pre-processing does not always guarantee Re-ID performance.

In order to deal with the low-illumination challenge in the nighttime person Re-ID, this work proposes an Illumination Distillation Framework (IDF), which utilizes illumination enhancement and illumination distillation schemes to promote the learning of the Re-ID model.
Specifically, IDF contains a master branch (MBranch), an illumination enhancement branch (IEBranch), and an illumination distillation module (IDModule).
MBranch is used to extract features from low-illumination nighttime images.
IEBranch comprises a self-supervised illumination enhancement network~\cite{DCE_2020} and a general feature extractor~(\emph{e.g.} ResNet50~\cite{he2016deep}).
The illumination enhancement network dynamically enhances each pixel of the nighttime image by combining multiple linear projections to generate an enhanced image.
The feature extractor extracts the features of the enhanced images.
Although there is a certain complementarity between nighttime and enhanced features, they have some limitations. 
For example, the quality of nighttime features is affected by low-illumination condition~\cite{ENGAN_2021}, while the enhanced features may introduce some inevitable noises due to overexposure or amplified noise in the extremely dark region~\cite{Ill_Invariant_2019ACMMM}.
Therefore, how to improve the complementary benefits of nighttime and enhanced features while suppressing data noises is a critical issue.
To this end, we design the IDModule, which fuses the features outputted from MBranch and IEBranch branches by the bottleneck fusion.
Therefore, we also apply distillation techniques between the fused features and the outputs of the MBranch and IEBranch to further improve the feature representation.
IDF is jointly optimizing the objective functions of illumination enhancement and re-identification tasks in an end-to-end learning scheme, which achieves an illumination enhancement approach dedicated to improving the performance of Re-ID tasks.  Thus, our IDF can better distinguish identities even in low-illumination nighttime scenarios.
%

It is worth discussing that the current mainstream approach to handle nighttime person Re-ID is to introduce infrared devices with better night vision capabilities to capture person images at night. However, due to the introduction of a new modality, resolving the large differences between infrared and visible modalities has become a major challenge in this field. For example, Hao~\emph{et al.}~\cite{CMMCCA_iccv} design a modality confusion learning network to eliminate the modality-awareness capability of the model, which facilitates the optimization of the model in the modality-invariance aspect. Unlike previous work based on the feature level, Ye~\emph{et al.}~\cite{vireid_trimodal_tifs} introduce intermediate modalities to mitigate modality differences from the image level and achieve significant performance improvement using a three-modalities learning approach. In addition, considering the noise problem of nighttime data annotation, Ye~\emph{et al.}~\cite{DTRM_tifs} simultaneously build channel-level, part-level intra-modality, and graph-level cross-modality relation cues to improve the issue of mismatching. Unlike existing works, the contribution of this paper is to propose a novel IDF approach to extend the existing Re-ID algorithm to nighttime scenes. In addition, this paper explores the potential of nighttime visible images and demonstrates that nighttime visible images are an alternative strategy to handle nighttime Re-ID.

To facilitate the study of nighttime person Re-ID, one existing work introduces a nighttime person Re-ID dataset~\cite{Zhang2019NightPR}~(\emph{i.e., Knight}) to the community.
However, this dataset is obtained with only three cameras and similar shooting angles, which makes it difficult to reflect the challenges in real-world nighttime environments.
In addition, 176 identities are captured under all three camera views in the \emph{Knight} dataset, but most identities are captured with only two cameras.
Considering these shortcomings of the \emph{Knight} dataset, this work contributes a new real-world nighttime person Re-ID dataset called \emph{Night600}. 
\emph{Night600} contains 600 person identities with a total of 28813 images.
These images are captured by eight non-overlapped cameras in nighttime scenes and cover different viewpoints with rich low-illumination conditions at night.

To sum up, this paper has the following major contributions.
\begin{itemize}

\item We propose a novel framework to address the challenge of low illumination in nighttime person Re-ID. It can utilize the complementary information from both low-illumination nighttime image and illumination-enhanced image while suppressing data noise through an illumination distillation module. 
\item We build a more challenging dataset to facilitate the research on nighttime person Re-ID. Compared with existing nighttime person Re-ID dataset, it contains more identities and camera shooting angles, and richer low-illumination conditions.
\item Extensive experiments on the proposed dataset and public dataset~\cite{Zhang2019NightPR} demonstrate that our framework performs favorably against the state-of-the-art methods.
\end{itemize}

\begin{figure*}[h]
    \centering
    \includegraphics[scale=0.45]{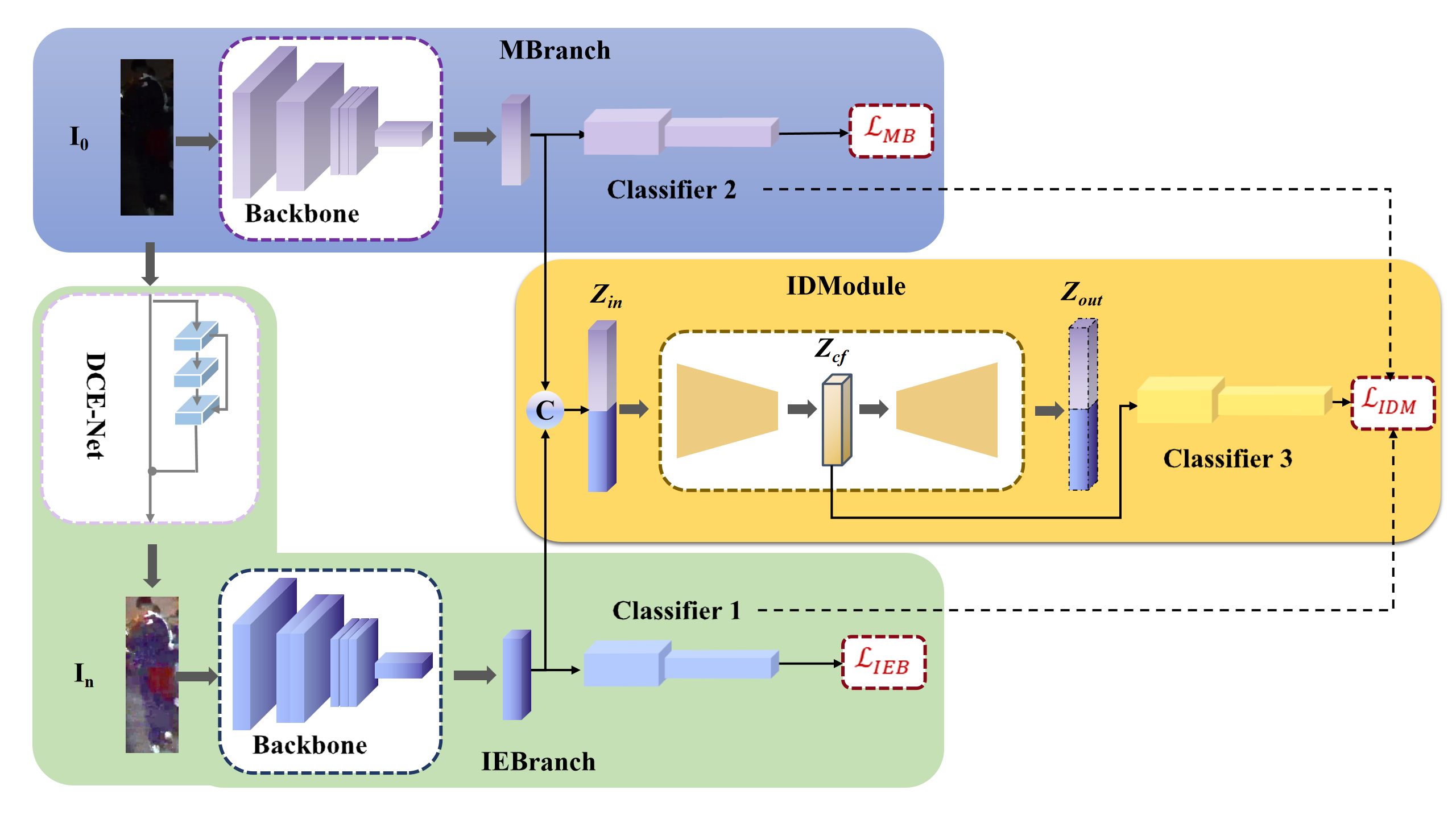}
    \vspace{-.5cm}
    \caption{Outline of the proposed framework, including master branch (MBranch), illumination enhancement branch (IEBranch), and illumination distillation module (IDModule). Herein, $\copyright$ represents the concatenation operation in the channel dimension.}
    \label{fig:model}
\end{figure*}

\section{Related Work}

\subsection{Person Re-ID Methods}

Most existing person Re-ID studies~\cite{PCB_2018,deng2018image,suh2018part,ABDNet_2019,li2018harmonious,song2018mask, HAT_2021,TransReID_2021,PCB_2018,RAGA_2020,CPA_tifs,AIIS_pami,CR_tip} focus on the learning of discriminative features 
through the guidance of elaborately designed classification or metric learning losses as well as the exploitation on human body structure priors.
%
For instance, some works propose different networks to learn discriminative features, including striping-based methods~\cite{PCB_2018,deng2018image,suh2018part} and local feature learning methods~\cite{ABDNet_2019,li2018harmonious,song2018mask, HAT_2021}.
These works aim at learning features from different body regions or try to align body parts between images.
There are also some works~\cite{TransReID_2021, PCB_2018, RAGA_2020} that combine local features and full-body features to obtain an enhanced identity representation. 
%
Some works~\cite{deng2018image,wang2018resource} adopt widely studied loss functions and their variants, including identity loss, verification loss, and triplet loss to improve model modeling capabilities.
To further improve recognition accuracy, Wu~\emph{et al.}~\cite{CPA_tifs} propose a novel multi-level context-aware part attention model to handle the challenges of severe occlusion, background clutter, and pose changes in video-based Re-ID, and it also designs multi-head collaborative training scheme to improve performance.
Ye\emph{et al.}~\cite{AIIS_pami} design two novel domain-agnostic augmentation strategies, and use data augmentation invariant and instance spreading feature to deal with the unsupervised embedding learning problem.
To address the label noise issue, Ye~\emph{et al.}~\cite{CR_tip} introduce an online co-refining framework with dynamic mutual learning that distills knowledge from other peer networks to further enhance robustness. 

Some special Re-ID tasks are also designed, such as black Re-ID~\cite{xu2020black}, occluded Re-ID~\cite{gao2020pose}, clothing change Re-ID~\cite{huang2019celebrities,gao2020pose}, and illumination change Re-ID~\cite{Real_world_CVPR2020, Ill_Invariant_2019ACMMM, IllAP_2020icip}. 
%
%
In black Re-ID~\cite{xu2020black}, only the hand-shoulder regions have strong discriminative information than other body parts due to all persons wearing similar black clothes.
Therefore, the module that is explicitly designed to extract hand-shoulder features is proposed for black Re-ID. 
However, the performance of the hand-shoulder feature extractor is affected by the robustness of pose estimators. 
In occluded Re-ID~\cite{gao2020pose}, the performance of models also heavily relies on the robustness of pose estimators due to the uncertainty of occluded regions.
In clothing change Re-ID~\cite{huang2019celebrities,gao2020pose}, the clothing information of persons becomes unreliable, while it is dominant in most traditional person Re-ID tasks.
In illumination change Re-ID, several works~\cite{Real_world_CVPR2020,Ill_Invariant_2019ACMMM,IllAP_2020icip} attempt to deal with the illumination change problem. 
These illumination change person Re-ID works rely on paired data (\emph{i.e.}, an RGB image and its corresponding synthesized low-illumination one) to learn illumination-invariant features  to deal with the low illumination issue in person Re-ID.
To this end, the gamma correction method~\cite{rahman2016adaptive} is used to synthesize low illumination images.
However, it is hard to simulate real low illumination conditions.
Moreover, although these methods are explicitly designed for illumination change person Re-ID and primarily on low-illumination challenges, they lack the capability of dealing with the challenges in real-world nighttime person Re-ID.

In this work, we focus on a more challenging nighttime person Re-ID problem, which is different from the above Re-ID problems. 
At night, a person appears in low-illumination conditions, which makes parts of the color and texture of the appearance invisible.
Although the night scene is even more important in certain circumstances than the daytime scene, this issue is rarely explored. 
We contribute a real nighttime person Re-ID dataset and propose a new IDF framework for the real low illumination issue in nighttime scenarios.

\subsection{Person Re-ID Datasets}
VIPeR~\cite{VIPeR} is the first and one of the most widely used person Re-ID dataset, including 632 identities and 1264 images captured by two cameras.
However, due to its limited training data scale, larger and richer datasets have been proposed, such as CUHK03~\cite{CUHK03}, Market1501~\cite{market1501_2015}, and DukeMTMC~\cite{DukeMTMC}, \emph{etc.}. 
These datasets are constructed with multi-camera views and multi-images for each identity.
In order to tackle more challenges in person Re-ID, huge scale person Re-ID datasets including Airport~\cite{Airport}, MSMT17~\cite{MSMT17}, and RPIfield~\cite{RPIfield} are introduced recently.
However, all the above-mentioned person Re-ID datasets are built based on daytime conditions.
The nighttime scenes which are more challenging do not get too much attention in existing literature. 

There is one existing dataset constructed at night, \emph{i.e., Knight}~\cite{Zhang2019NightPR}.
It contains 971 identities and 315,354 images captured by three camera views.
However, \emph{Knight} uses only a limited number of camera views with similar shooting angles (as shown in Fig.~\ref{fig:examples_fig}~(b)). 
It is not able to reflect the real low illumination condition at night since the illumination is normally quite low and unstable in the nighttime.
In addition, there are only 176 identities captured by all three camera views, and a majority of identities only appear under two views.
In this work, we propose a new nighttime person Re-ID dataset that takes more camera shooting angles with richer low-illumination conditions into consideration. 
\subsection{Knowledge Distillation Methods}

Traditional knowledge distillation methods~\cite{hinton2015distilling,huang2017like,li2020few,heo2019knowledge} are usually offline, which transfer a pre-trained large-scale teacher model to a small-scale student model. Current offline methods focus on different aspects of knowledge transfer, such as designing knowledge~\cite{hinton2015distilling} and different loss functions~\cite{huang2017like,li2020few,heo2019knowledge}. Although these methods are simple and easy to implement, it requires a complex high-capacity teacher model, which costs a huge training time. To overcome the limitation of offline distillation, existing distillation techniques utilize different strategies to achieve online knowledge transfer between multiple models, which greatly facilitates the application of the technique in practical tasks. In the online distillation scheme, the teacher and student models are updated simultaneously in an end-to-end framework. For instance, Guo~\emph{et al.}~\cite{guo2020online} treat all networks as students and collaboratively train them in one stage to achieve knowledge transfer among arbitrary students. Chung \emph{et al.}~\cite{chung2020feature} design an online adversarial knowledge distillation method that uses discriminators to guide category probability and feature map distillation. Inspired by the one teacher vs. multiple students pattern in schools, Shen \emph{et al.}~\cite{shen2021distilled} propose transferring knowledge from the teacher model to the student model, and between student models, which yields effective distillation results in the tracking task. In contrast to these approaches, we pursue learning different student models and then aggregating them into a teacher model.

%
\section{Illumination Distillation Framework}
In this section, we will describe the Illumination Distillation Framework (IDF) in detail, including the overview of network architecture, the illumination enhancement branch, the master branch, the illumination distillation module, and the training and inference details.

\subsection{Overview of Network Architecture}
The proposed network architecture is shown in Fig.~\ref{fig:model}, which contains a master branch (MBranch), an illumination enhancement branch (IEBranch), and an illumination distillation module (IDModule).   
Given a nighttime person image $I_{0}$ as the input, we simultaneously feed it into MBranch and IEBranch.
In IEBranch, $I_{0}$ first passes the illumination enhancement network DCE-Net~\cite{DCE_2020} to obtain its illumination-enhanced image.
The illumination-enhanced image of $I_{0}$ is then fed into a feature extractor to obtain the enhanced features, where the feature extractor is replaceable with any off-the-self Re-ID model (\emph{e.g.,} IDE~\cite{IDE+_2019}, AGW~\cite{AGW_2021}, TransReID~\cite{TransReID_2021}, etc.).
In MBranch, we directly extract the nighttime features from $I_{0}$.
After that, the nighttime features and enhanced features are concatenated, and the concatenated features are sent to IDModule to explore the complementary benefits of the two branches.
In order to promote the effectiveness of both branches, we take IDModule as a role of teacher to further improve the robustness of features learned from MBranch and IEBranch, respectively.
Finally, all features learned from the two branches and IDModule are fed into individual classifiers to differentiate different identities.

\subsection{Illumination Enhancement Branch}
Since persons usually appear in low illumination conditions in the nighttime, which induces low contrast between foreground and background regions.
Therefore, how to extract the identity features from person images under low illumination conditions is the main concern in nighttime person Re-ID.
Although many technologies~\cite{DCE_2020,ENGAN_2021} try to enhance the contrast of nighttime images, they mainly focus on improving subjective visual quality rather than improving the discriminability of different subjects presented in different images.
%
%
To handle this issue, we design an IEBranch to learn discriminative features for nighttime person Re-ID. 
IEBranch contains a self-supervised illumination enhancement network DCE-Net~\cite{DCE_2020} and a general features extractor (\emph{e.g.} ResNet50~\cite{he2016deep}). 
It first uses DCE-Net to enhance the nighttime image, and then the feature extractor is employed to extract enhanced features from the illumination-enhanced image.  
Specifically, nighttime image $I_{0}$ first inputs to DCE-Net, which is composed of seven convolutional layers with symmetrical skip concatenation, and then outputs a set of pixel-wise non-linear curve parameter maps $\mathcal{A}$ with the same size as the input image.
Then, each pixel value of $I_{0}$ is adjusted to an appropriate range according to $\mathcal{A}$ with several iterations, which can be calculated as:
\begin{equation}
\begin{aligned}
&{LE(I_{0},\mathcal{A})= {I_0}+{\mathcal{A}*I_0}\ast{(1-I_0)}}\\
&{I_{n} = {LE(I_{n-1};\mathcal{A}_n)}},\\
\end{aligned}
\label{eq::1}
\end{equation}
where $LE$ represents the enhancement function, $I_n$ is the enhanced person image in the $n$-th iterations, and following ~\cite{DCE_2020} we set the number of iterations to 8.

DCE-Net adopts a self-supervised learning fashion to enhance the illumination of an image through four self-supervised losses in the training stage, which is practical to real-world nighttime scenarios.
In detail, DCE-Net applies the spatial consistency loss $\mathcal{L}_{spa}$ to encourage spatial coherence of the illumination-enhanced image by keeping the difference of local regions between the nighttime image and its illumination-enhanced image:
\begin{equation}
\begin{aligned}
&\mathcal{L}_{s p a}=\frac{1}{M} \sum_{m=1}^{M} \sum_{k \in \Omega(m)}\left(\left|\left(I_{n}^{m}-I_{n}^{k}\right)\right|-\left|\left(I_{0}^{m}-I_{0}^{k}\right)\right|\right)^{2},\\
\end{aligned}
\label{eq::2}
\end{equation}
where $M$ is the number of local regions, and $\Omega(m)$ represents four neighboring regions centered on the region $m$. We set the size of the local region to $4 \times 4$ followed by~\cite{DCE_2020}.

The exposure control loss $\mathcal{L}_{exp}$ is used to control the exposure level by measuring the distance between the average intensity value of the illumination-enhanced image to the well-exposed constant (\emph{i.e.}, 0.6), which can be expressed as:
\begin{equation}
\begin{aligned}
&\mathcal{L}_{e x p} = \frac{1}{K} \sum_{k = 1}^{K}\left|I_{n}^{k}-0.6\right|,\\
\end{aligned}
\label{eq::3}
\end{equation}
where $K$ represents the number of non-overlapping local regions of the size $16 \times 16$.

The illumination smoothness loss $\mathcal{L}_{tv_A}$ is used to avoid producing too much noise on the illumination-enhanced image, which is expressed as follows:
\begin{equation}
\begin{aligned}
&\mathcal{L}_{t v_{\mathcal{A}}}=\frac{1}{N} \sum_{n=1}^{N} \sum_{c \in \xi}\left(\left|\nabla_{x} \mathcal{A}_{n}^{c}\right|+\left|\nabla_{y} \mathcal{A}_{n}^{c}\right|\right)^{2}, \xi=\{R, G, B\},\\
\end{aligned}
\label{eq::4}
\end{equation}
where $N$ is the number of iteration, and $c$ represents one of the channels of the illumination-enhanced image. 
The $\nabla_{x}$ and $\nabla_{y}$ represent horizontal and vertical gradient operations, respectively.

The color constancy loss $\mathcal{L}_{col}$ is applied to control the color shift problem in the process of enhancement through the Gray-world color constancy hypothesis~\cite{chollet2017xception}, which can be described as follows: 
\begin{equation}
\begin{aligned}
&\mathcal{L}_{\mathrm{col}} =\sum_{\forall(i,j) \in \varepsilon}\left(C^{i}-C^{j}\right)^{2}, \varepsilon=\{(R, G),(R, B),(G, B)\},\\
\end{aligned}
\label{eq::5}
\end{equation}
where $C^{i}$ indicates the average intensity value of the $i$-th channel of the illumination-enhanced image.

Finally, the combination of four self-supervised losses for illumination enhancement $L_{DCE}$ is defined by:
\begin{equation}
\begin{aligned}
&L_{DCE}= \mathcal{L}_{spa} + \mathcal{L}_{exp} + \mathcal{L}_{tv_{\mathcal{A}}} + \mathcal{L}_{\mathrm{col}},\\
\end{aligned}
\label{eq::6}
\end{equation}
Note that the weights of these losses are same as~\cite{DCE_2020}. 
Moreover, the purpose of our method is to improve the performance of person Re-ID rather than generate visually pleasing images. 
Therefore, DCE-Net in our architecture trained with the Re-ID task through minimizing ID loss $L_{ID}$ to classify different subjects.
Therefore, the process of illumination enhancement can benefit the semantic consistency between the original low-illumination image and the corresponding illumination-enhanced image produced by DCE-Net.
Specifically, we minimize the difference between the predicted probability of identity and the ground-truth identity as follows:
\begin{equation}
\label{eq.7}
\begin{aligned}
&L_{ID}=-\frac{1}{B} \sum_{b=1}^{B} \log{\tilde{y}}_{p}^{b},
\end{aligned}
\end{equation}
where $B$ is the number of images in a mini-batch, and $\tilde{y}_{p}$ is the predicted probability of the illumination-enhanced image belonging to the ground-truth ID label. 
Therefore, the overall loss of the illumination enhancement branch is defined by:
\begin{equation}
\begin{aligned}
&L_{IEB}= L_{ID} + L_{DCE}.
\end{aligned}
\label{eq::8}
\end{equation}

\subsection{Master Branch}
Although the illumination-enhanced image may be visually better than the nighttime image, some noises may be introduced during the illumination enhancement process, especially on very dark regions~\cite{Ill_Invariant_2019ACMMM}.
Therefore, we build a MBranch to capture additional information from nighttime images, which can be regarded as complementary to the IEBranch.
It could mitigate the interference of noise or over-enhancement in illumination-enhanced images.
The feature extractor of this branch is same as IEBranch for simplicity, which also can be replaced with other Re-ID networks. 
%
MBranch is driven by the ID loss to directly classify nighttime person images, and thus the loss of MBranch $L_{MB}$ defined by:
\begin{equation}
\label{eq.7}
\begin{aligned}
&L_{MB} = L_{ID}=-\frac{1}{B} \sum_{b=1}^{B} \log{\tilde{y}}_{p}^{b},
\end{aligned}
\end{equation}
Note that we do not share parameters between MBranch and IEBranch to facilitate the modeling of different illumination conditions from input images. 

%

\subsection{ Illumination Distillation Module} 
To handle complex low illumination challenge in nighttime person Re-ID, a straightforward idea is to fuse the outputted features of MBranch and IEBranch.
However, the quality of nighttime features from MBranch is unstable under different illumination conditions, and the enhanced features from IEBranch may also contain noise due to over-enhancement.
Therefore, how to achieve effective complementary fusion between nighttime and enhanced features while suppressing data noises is critical. 

To this end, we design an illumination distillation module (IDModule), which contains a bottleneck fusion model and a classifier.
Specifically, the bottleneck fusion model, which consists of an encoder and a decoder, is used to fuse the features of MBranch and IEBranch while suppressing feature noise.
The encoder and the decoder have a symmetrical structure, which includes two fully-connected layers and a nonlinear activation function $ReLU$.
The classifier is composed of two fully-connected layers in which a batch norm layer and a dropout layer are followed after the first fully-connected layer.  
We denote the inputted features of the bottleneck fusion model as $z_{in}$, which is obtained by concatenating the output features of MBranch and IEBranch.
Then $z_{in}$ is fed into the encoder to obtain compact fused features ${z}_{cf}$.
After that, we respectively feed $z_{cf}$ into the decoder and the classifier for reconstruction and classification.

As shown in Fig.~\ref{fig:model}, the bottleneck of encoder-decoder can reduce the dimension of $z_{in}$, which promotes the model to eliminate a certain amount of information irrelevant to reconstruction and classification tasks. 
The reconstruction task is used to guide the compact features~(\emph{i.e.}, $z_{cf}$) to preserve the more comprehensive information in $z_{in}$ as much as possible.
Here, we minimize the Euclidean distance between $z_{in}$ and reconstructed features $z_{out}$ as the reconstruction loss, as follows:
\begin{equation}
\begin{aligned}
&{L_{REC} = \left\|z_{out}-z_{in} \right\|^{2}}.
\end{aligned}
\label{eq::9}
\end{equation}
Moreover, $z_{cf}$ is also forwarded to the classifier for the classification task based on ID loss, which ensures the discriminability of $z_{cf}$.

To further refine the learning of MBranch and IEBranch, we propose a distillation strategy for nighttime person Re-ID.
Unlike traditional two-stage offline distillation scheme that first trains a teacher model and then distillates the knowledge from the teacher model to a student model~\cite{KD_tits_2}, the one-stage online distillation scheme shows large progress in existing literature~\cite{onlinekl_2021iccv, shen2021distilled, KD_tits_1}. The one-stage online distillation scheme can establish a teacher model on-the-fly.
For instance, Shen \emph{et al.}~\cite{shen2021distilled} selects several small networks as student models from a large-capacity teacher model and performs knowledge transfer between them to obtain a high-performance lightweight network.
However, our approach aggregates the output features of two student models (IEBranch, MBranch) with a specially designed fusion module and then applies a distillation technique based on the category probability distribution to transfer knowledge.
Furthermore, ~\cite{shen2021distilled} requires enough training data to learn a good large-capacity teacher model, while the data scale of nighttime scenarios is often limited. However, our approach requires learning only two small-capacity student models, which is easier to achieve under the constraints of small-scale data.

Following the above scheme, we take the bottleneck fusion model as our teacher model and the other two branches as the student models in the training stage.
To transfer the identity-related knowledge from the teacher model to the student models, we employ the Kullback-Leibler (KL) divergence loss to minimize the divergence between the output of the teacher model and the outputs of the two student models as follows:

\begin{equation}
\begin{aligned}
&{L_{IFD} = - \sum_{t \in T} \sum_{s \in S}\sum_{n=1}^{N}P_{t}(n)\log\frac{P_{s}(n)}{P_{t}(n)} },\\
\end{aligned}
\label{eq::10}  
\end{equation}
where $P_{t}^{n}$ and $P_{s}^{n}$ are the probabilities of identity predicted by the teacher model and the student model, respectively. 
$S$ and $T$ represent the set of student model (MBranch, IEBranch) and the set of teacher model(IDM), respectively. 
This design enables the two branches to learn more effective identity discriminative knowledge from each other, which further facilitates the learning of both branches.
Finally, the overall loss of illumination distillation module is defined by:
\begin{equation}
\begin{aligned}
&L_{IDM}= L_{ID} + \lambda_1 L_{REC} + \lambda_2 L_{IFD} ,
\end{aligned}
\label{eq::11}
\end{equation}
where $\lambda_1$ and $\lambda_2$ are the balancing factors, which are set to 0.1 after experimental verification.

\subsection{Training and Inference}
The whole network is trained in an end-to-end manner and the overall objective function is expressed as:
\begin{equation}
\begin{aligned}
&\underset{\theta_{mb}, \theta_{ieb}, \theta_{idm}}{\arg \min }~~ L_{IEB}+L_{MB}+L_{IDM}, \\
\end{aligned}
\label{eq::12}
\end{equation}
where $\theta_{mb}$, $\theta_{ieb}$ and $\theta_{idm}$ represent the parameters of MBranch, IEBranch and IDModule, respectively.
In the training stage, we employ Stochastic Gradient Descent (SGD) with momentum of 0.9 and weight decay of 0.0005 for the optimization of the overall objective function.
MBranch and IEBranch encourage our model to handle different illumination conditions in nighttime scenarios.
IDModule enables our model to fuse the features of low illumination image and enhanced image, and guides both branches to be mutually complementary between each other.  
Moreover, the joint learning manner can better balance the tasks of illumination enhancement and nighttime person Re-ID.

During the inference, we use the concatenated features from the outputted features of MBranch, IEBranch, and the encoder of IDM to perform similarity ranking to carry out nighttime person Re-ID.

\section{Night600: Nighttime Person Re-ID Dataset}

\begin{table*}[]
 \centering
\caption{Comparison of \emph{Night600} with other datasets.}
\label{tab:dataset_info}
\begin{tabular}{c|cc|cccccc}
\hline
\multirow{2}{*}{Dataset} & \multicolumn{2}{c|}{Nighttime Data} & \multicolumn{6}{c}{Daytime data}                                 \\ \cline{2-9} 
                & Night600 & Knight  & i-LIDS & VIPeR & CUHK01 & CUHK03 & Market1501 & MARS      \\ \hline
Identity Number & 600      & 937     & 119    & 632   & 971    & 1467   & 1501       & 1261      \\
Camera Number   & 8        & 3       & 2      & 2     & 2      & 6      & 6          & 6         \\
Image Number    & 28,813   & 315,354 & 476    & 1,264 & 3,884  & 14,097 & 32,217     & 1,119,003 \\
Viewpoint                & Parallel/Downward     & Downward    & Downward & Parallel & Downward & Downward & Parallel & Downward  \\
Labeling Method          & Detectorn2/Hand       & Hand        & Hand     & Hand     & Hand     & DPM/Hand & Hand     & DPM/GMMCP \\ 
Glimmer Image       & YES       & NO       & NO     & NO     & NO     & NO & NO    & NO \\ \hline
\end{tabular}
\end{table*}

\begin{figure}
\centering
\subfloat[]{
\begin{minipage}[b]{0.42\textwidth}
  \includegraphics[width=1\textwidth]{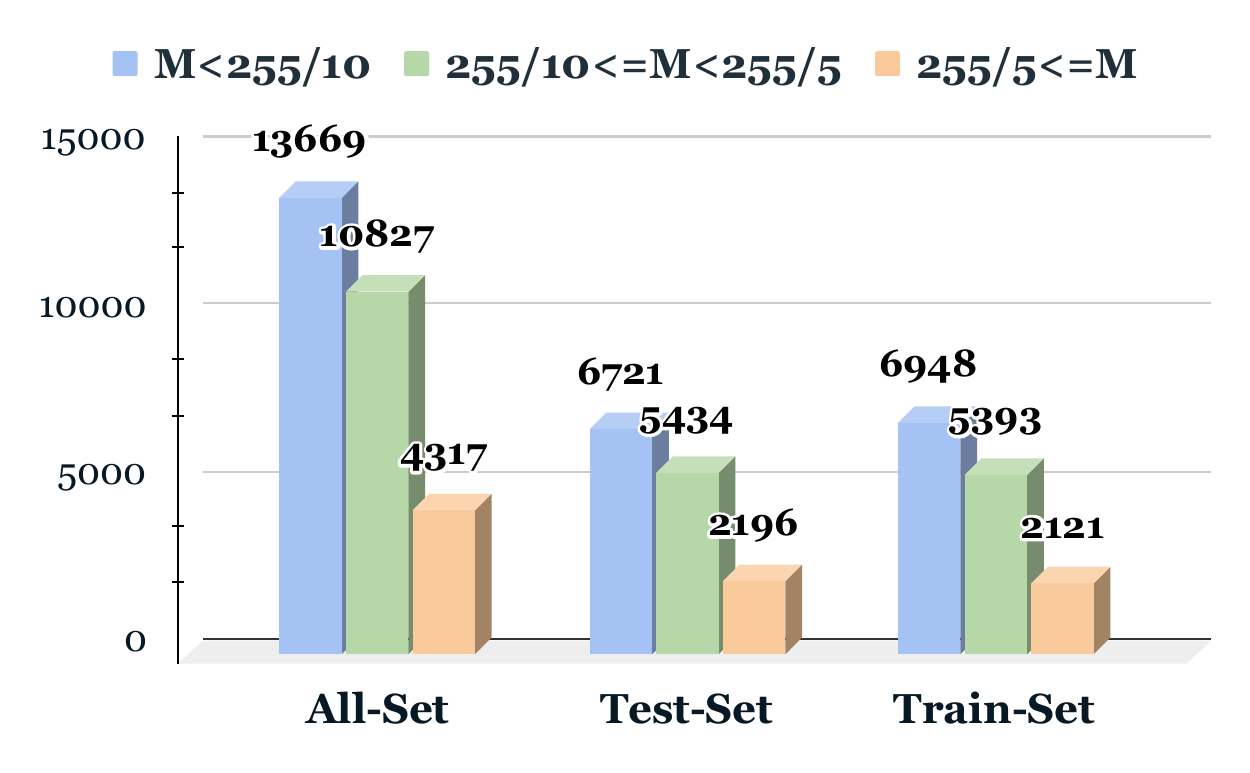}
  \end{minipage}
}
\\
\subfloat[]{
\begin{minipage}[b]{0.42\textwidth}
  \includegraphics[width=1\textwidth]{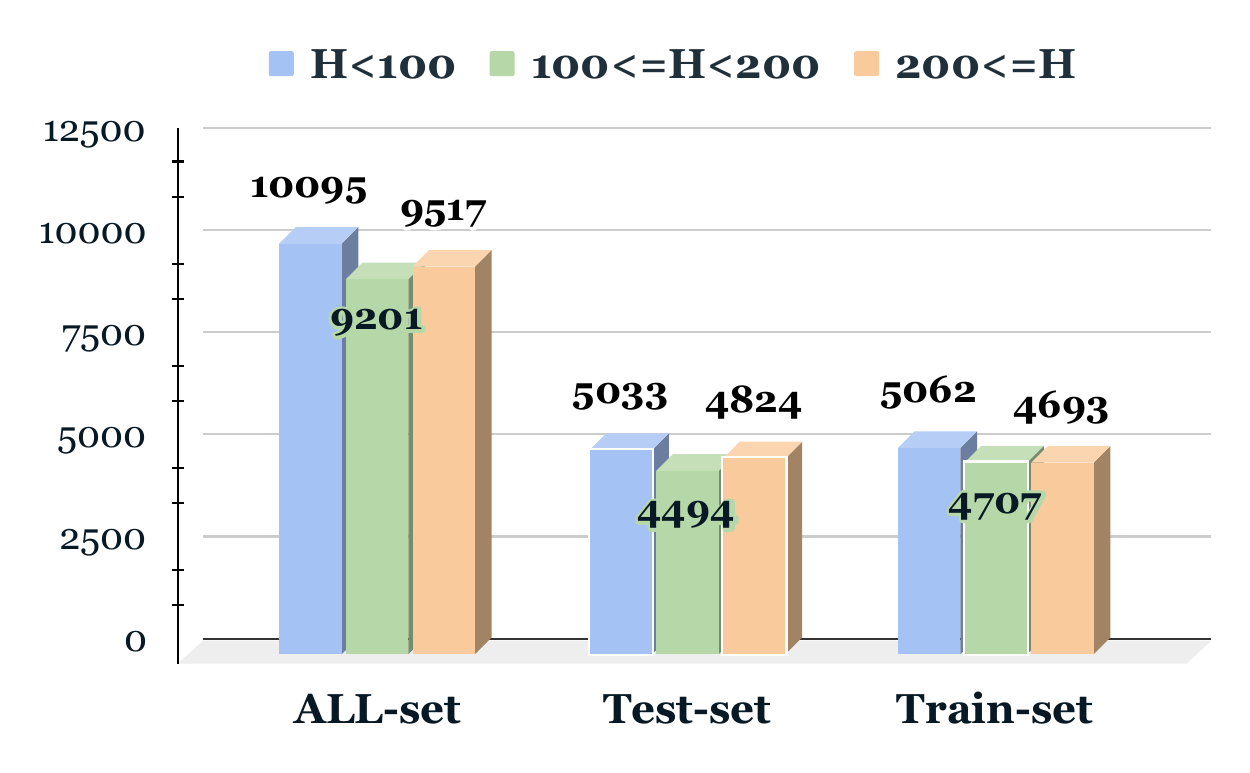}
  \end{minipage}
}
\caption{(a) and (b) show the statistical results of the illumination and scale distributions in \emph{Night600} dataset, respectively. Herein, $H$ represents the height of the image, $M$ denotes the mean lightness of images. }
\label{fig:statistics_dataset}
\end{figure}

To boost the studies of person Re-ID in the nighttime, a real-world nighttime dataset \emph{Night600} with diverse viewpoints and illumination conditions, is collected as a new benchmark to the community. 

\subsection{Dataset Description}
\noindent\textbf {Data Acquisition. } The \emph{Night600} dataset is collected from real-world nighttime scenes using eight non-overlapping cameras, which have parallel and downward viewpoints in common surveillance scenarios.
Specifically, we utilize eight visible-light cameras with a resolution of 1920$\times$1080  to capture images of people on campus roads during the night.
%
Totally 3.17G video data are collected.
Note that, the real-world nighttime videos cover complex situations with various illuminations, such as street lamps, vehicle headlights, advertisement board reflective lights, and starlight, \textit{etc.}, which are very challenging.

\noindent\textbf{Data annotation.} The whole annotation process consists of three stages as follows. 
1) We convert the video data to image frames, and select a total of about 22000 frames containing persons from the captured video data.
2) We ensure that each identity is captured by at least two cameras to satisfy the need for cross-camera person retrieval.  
Moreover, we also annotate some characteristic persons, such as running, cycling, and taking bags.
3) To alleviate the costs of manual annotations, we utilize a semi-automatic annotation method to collect image samples.
Specifically, we use detectron2~\cite{wu2019detectron2} to obtain initially detected boxes of all frames, and then manually annotate undetected persons while correcting the wrong detected boxes.
Here, the manually annotated boxes can provide more samples of person within glimmer conditions, which is more challenging for nighttime person Re-ID. 

\noindent\textbf{Data Statistics.} As shown in Table~\ref{tab:dataset_info}, \emph{Night600} dataset contains 28,813 images of 600 identities with eight camera viewpoints. 
We choose 300 identities for training and 300 identities for testing. 
Then, we randomly select 3 images from each viewpoint of each identity as probes, and the others as the gallery set. 
Finally, 2180 probes are obtained as the query set, 14,462 images as the training set, and 14,351 images as the gallery set.

Moreover, the dataset is divided into different partitions in terms of various illuminations and image scales, which can be used to investigate the properties of Re-ID models under different environmental factors. 
Specifically, the illumination conditions of city roads at night can be divided into unnatural light (street lamp, vehicle light, and electric light device) and natural light (starlight or moonlight) illumination conditions. 
According to the histogram of the image illumination, we divide person images into three illumination levels: Low ($<255/10$), Medium ($255/10-255/5$), and High ($>255/5$). 
According to the height of the image, we also divide person images into three different scale levels: Small ($<100$), Medium ($100-200$), and Big ($>200$). 
The majority of this dataset contains glimmer images, because of the local lighting in the nighttime.
In addition, this dataset contains diverse scales caused by the multiple camera views.
In (a) and (b) of Fig.~\ref{fig:statistics_dataset}, we show the illumination and scale distributions in the training, testing, and entire sets.

\subsection{Challenges}
\begin{figure}[h]
    \centering
    \includegraphics[scale=0.25]{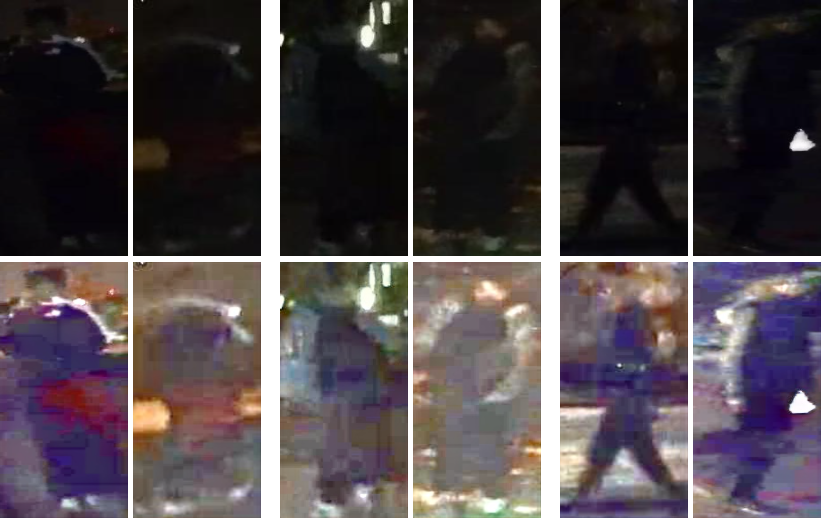}
    \vspace{-.3cm}
    \caption{Some examples of glimmer person images from \emph{Night600} dataset. The first row is the original person images, and the second row is the corresponding image enhanced by low-illumination enhancement method~\cite{ENGAN_2021}}
    \label{fig:Data_example}
\end{figure}

In nighttime person Re-ID, the most challenging problem is glimmer person images, which are almost invisible to human eyes as shown in the first row of Fig.~\ref{fig:Data_example}.
Although existing low-illumination enhancement methods enable to enhance the illumination of these images, it inevitably introduces more noise, as shown in the second row of Fig.~\ref{fig:Data_example}.
Therefore, how to effectively extract more robust features from glimmer images and their illumination-enhanced images is still a challenging task in nighttime person Re-ID.

\begin{figure}[h]
    \centering
    \includegraphics[scale=0.42]{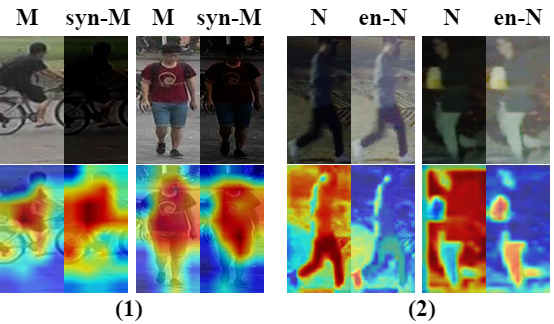}
    \vspace{-.3cm}
    \caption{Classification activation maps of AGW on the synthetic low-illumination Market-1501 (syn-M) dataset, the traditional daytime Market-1501 (M) dataset, the proposed \emph{Night600} dataset (N), and enlightened \emph{Night600} dataset (en-N), respectively.}
    \label{fig:CAM}
\end{figure}

Moreover, we conduct a set of visualization experiments to better show why existing Re-ID algorithms are hard to process nighttime Re-ID.
We train AGW~\cite{AGW_2021} on the daytime dataset \emph{Market1501}~\cite{market1501_2015}, synthetic low-illumination dataset \emph{syn-M}, and nighttime dataset \emph{Night600} and illumination-enhance nighttime dataset \emph{en-N} in experiments.
%
%
From Fig.~\ref{fig:CAM} (1), it can be seen that the synthetic dataset is hard to simulate real-world nighttime scenarios, as the AGW activation maps of synthetic nighttime images are still relatively similar to their corresponding daytime images.
%
%
Fig.~\ref{fig:CAM} (2) explains why the poor performance of AGW in nighttime dataset.
The main reason is that the model cannot well distinguish foreground and background information.
Moreover, it can be seen that directly using illumination enhancement techniques to pre-process nighttime data only achieves sub-optimal performance.

\subsection{Difference of \emph{$Night600$} from \emph{$Knight$}}
\begin{figure*}[h]
    \centering
    \includegraphics[scale=0.24]{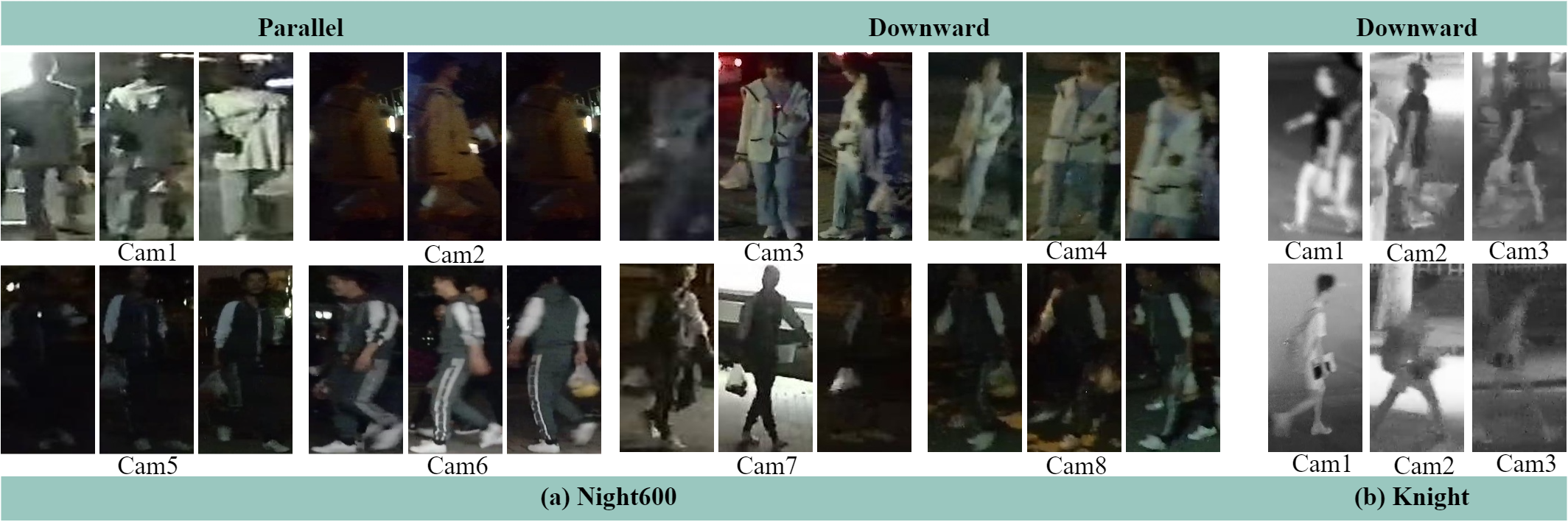}
    \vspace{-.3cm}
    \caption{Some typical examples of \emph{Night600} and \emph{Knight} datasets. \emph{Night600} shows more diverse illumination conditions and camera views than \emph{Knight}.}
    \label{fig:examples_fig}
\end{figure*}

\begin{figure}[h]
    \centering
    \includegraphics[scale=0.11]{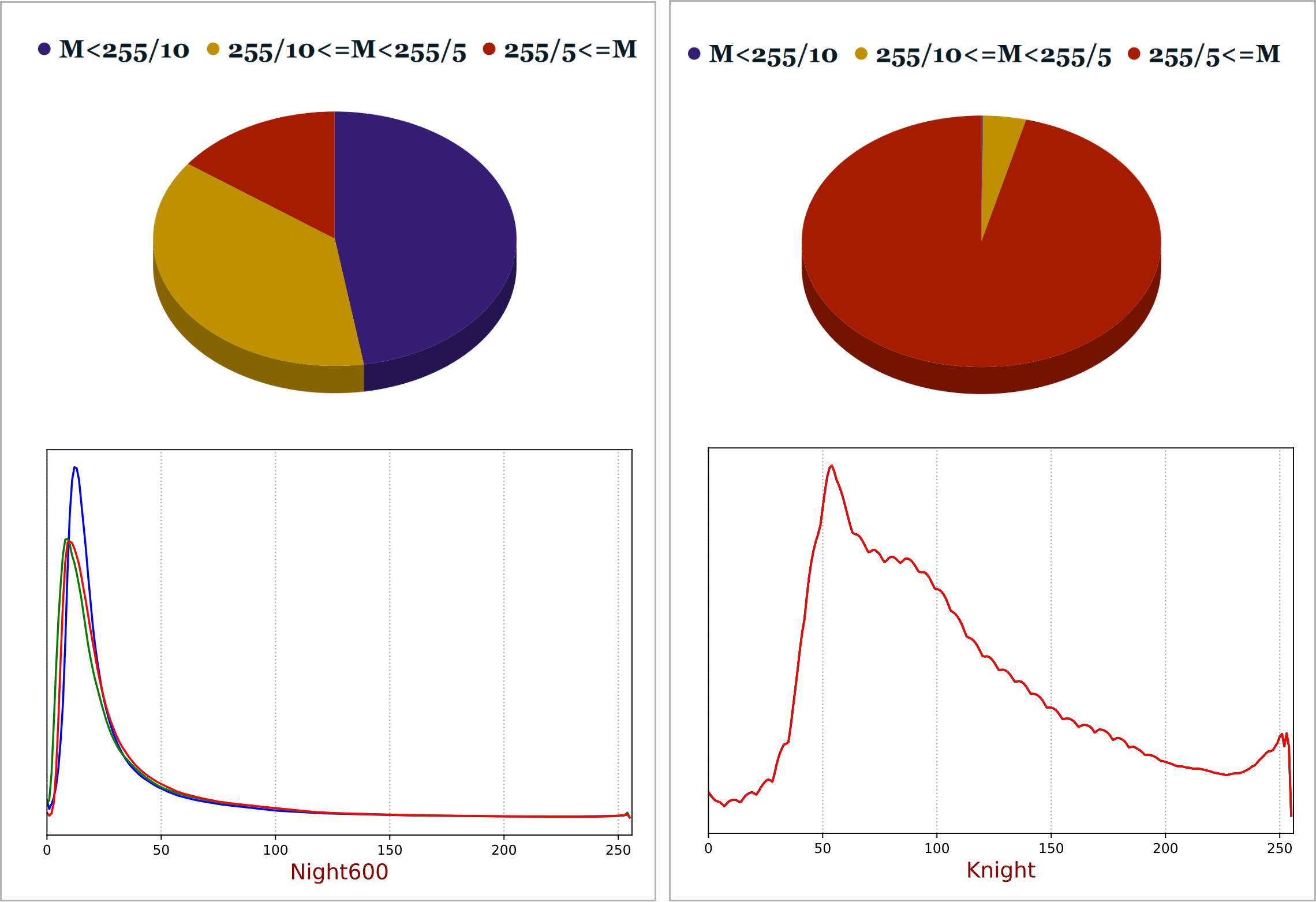}
    \caption{Some typical statistics of \emph{Night600} and \emph{Knight} datasets. Our dataset has more diverse illumination conditions and camera views.}
    \label{fig:comapre_dataset_fig}
\end{figure}

Person images usually suffer from low illumination condition in the nighttime, which is a key challenge of nighttime person Re-ID.
%
As shown in Fig.~\ref{fig:examples_fig}, we select the samples of several representative identities from \emph{Night600} and \emph{Knight}~\cite{Zhang2019NightPR} to show some differences between these two datasets.
It can be seen that the illumination conditions and camera viewpoints are the main differences.
The criterion to evaluate a dataset is whether it covers the variations in real scenes as much as possible, and thus we employ eight cameras including different viewpoints (parallel and downward) to capture person images in different nighttime scenes.
Compared with our dataset, there are only three cameras with similar viewpoints in \emph{Knight} dataset, which make it lack the capability to evaluate person Re-ID models in complex scenarios.
Moreover, the person images captured under glimmer lighting conditions, \emph{e.g.}, $cam2$, $cam5$ and $cam8$, pose more challenging in nighttime scenarios.
For a clear comparison, we show the statistical distributions of images under three illumination levels and histograms of the R, G, and B channels in \emph{Knight} and \emph{Night600} datasets. 
The results are presented in Fig.~\ref{fig:comapre_dataset_fig}.
From the statistical results in the first row, it can be seen that our \emph{Night600} dataset provides more diverse and balanced data distributions than the \emph{Knight} dataset in low-illumination nighttime scenarios. 
From the distribution of data in \emph{Knight} dataset, the corresponding pixel values in R, G, and B channels are too close which cannot reflect the real color information shown in the nighttime.

Up to now, there still lacks a large-scale nighttime Re-ID benchmark collected in the real world. 
Compared to the one existing real-world nighttime Re-ID dataset (\emph{i.e.} \emph{Knight}), the proposed \emph{Night600} provides more diverse camera viewpoints and illumination conditions for comprehensive evaluations of nighttime Re-ID algorithms.
Although \emph{Night600} is not a large-scale Re-ID dataset, the scales of existing nighttime Re-ID datasets are enough to the learning of CNN-based Re-ID models. 
As shown in Table~\ref{tab:sota_compare_exp} of the experiments section, our method achieves obvious improvement on CNN-based models (IDE+ and AGW), which indicates that the proposed model can be sufficiently learned on the scale of existing nighttime datasets. 
In the future, we will further expand this dataset for the training of Transformer-based models.

\subsection{Necessity Analysis of Visible Image based Nighttime Person Re-ID}
Herein, we discuss why nighttime Re-ID tasks are explored based on visible images rather than infrared ones.
Although the infrared images seem more reasonable for the task of nighttime Re-ID, there are three major reasons to study visible image based nighttime Re-ID as follows.
First, despite infrared cameras can capture clear person images at low illumination conditions, visible cameras are the majority in current surveillance systems. 
Therefore, the study of RGB-based nighttime Re-ID is essential.
Second, due to the limitation of the infrared light spectrum, infrared images have intrinsic flaws to capture the detailed color and texture information, which plays a critical role in Re-ID task. 
Third, the training of deep learning models relies on large-scale pre-trained models which are mostly trained on visible image datasets, {\emph{e.g.} ImageNet}. 
The domain gap between visible and infrared images hinders the training of infrared image based person Re-ID models. 
Thus, it is necessary to study visible image based nighttime person Re-ID.
Finally, recent advances in image illumination enhancement~\cite{DCE_2020,jiang2022unsupervised,ENGAN_2021} have proved the effectiveness of relighting visible images in low illumination scenes, and vision tasks for nighttime scenes are developing rapidly in other fields such as detection~\cite{hassaballah2020vehicle} and segmentation~\cite{wang2022sfnet}. 
It motivates us to explore the task of visible image based nighttime Re-ID. 

\begin{figure}[h]
    \centering
    \includegraphics[scale=0.4]{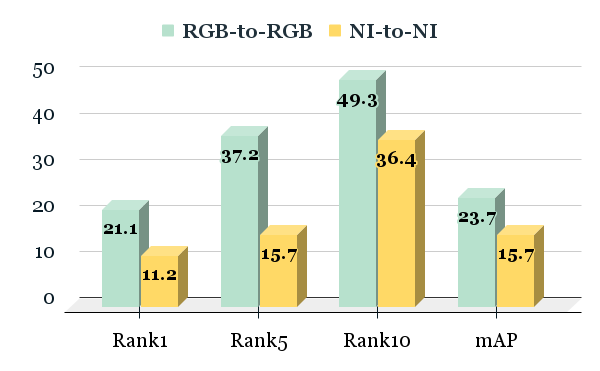}
    \vspace{-.3cm}
    \caption{Evaluation results of the AGW~\cite{AGW_2021} algorithm for the visible modality subset (RGB-to-RGB) and infrared modality subset (NI-to-NI) in the RGBNT201~\cite{RGBNT201_aaai} dataset.}
    \label{fig:agw_rgbt201}
\end{figure}

In addition, we introduce the RGBNT201~\cite{RGBNT201_aaai} dataset to validate the performance of visible and infrared modalities in the nighttime scene Re-ID task. This dataset is a multi-modality person Re-ID dataset, which contains highly aligned images of people in visible and infrared modalities at night. We use AGW~\cite{AGW_2021} to perform re-identification in the same modality. As shown in Fig.~\ref{fig:agw_rgbt201}, it can be seen that the AGW~\cite{AGW_2021} achieves the best performance in the visible modality, indicating that the visible modality still has advantages even in nighttime scenarios.

\section{Experiment}
In this section, we first introduce the experimental settings, then perform an overall evaluation and comparison on two nighttime person Re-ID datasets, and lastly conduct an in-depth analysis to verify the effectiveness of our method.  

\subsection{Experimental Settings}
\noindent{\bf Implementation Details.} The implementation platform of the proposed approach is Pytorch. 
All the used images are resized to $256\times128\times3$.
We use three representative and high performance person Re-ID models as the feature extract of MBranch and IEBranch, including IDE+~\cite{IDE+_2019}, AGW~\cite{AGW_2021}, and TransReID~\cite{TransReID_2021}.
For IDF$_{IDE+}$ we use ResNet50~\cite{he2016deep} pre-trained on ImageNet~\cite{deng2009imagenet} as the backbone. 
The initial learning rate, number of iterations, and batch size are respectively set to 0.0005, 120, and 32, which follows the setting in~\cite{IDE+_2019}.
For IDF$_{AGW}$, the ResNet50~\cite{he2016deep} with non-local attention is adopted as the backbone, and the number of iterations is set to 120, and other settings are the same as AGW~\cite{AGW_2021}. 
For IDF$_{TransReID}$, we use a vision transformer~\cite{ViT_2021} pre-trained on ImageNet~\cite{deng2009imagenet} as the backbone, and the number of iterations is set to 300, and other settings are the same as TransReID~\cite{TransReID_2021}.
\\

\noindent{\bf Datasets.} We conduct experiments on two nighttime Re-ID datasets: \emph{Night600} and \emph{Knight}~\cite{Zhang2019NightPR}.
For \emph{Night600}, it consists of 14,462 images of 300 identities for training and 14,351 images of 300 identities in gallery for testing.
For \emph{Knight}, it divides the three cameras into three subsets for training and testing. 
In detail, cam1-cam2 contains 183,430 images of 474 identities, cam1-cam3 contains 56,889 images of 176 identities, and cam2-cam3 contains 153,547 images of 639 identities.
All the above three settings follow an evaluation protocol in that half of the identities are used for training and the other half for testing.

\begin{table}[h]
\centering
\caption{Performance of existing advanced Re-ID methods on the \emph{Night600} dataset.}
\label{tab:sota_compare_exp}
\begin{tabular}{cccccc}
\hline
                     & R-1  & R-5   & R-10  & $m$AP  \\\hline
BoT            & 11.20 & 22.60 & 30.60 & 5.40 \\
ABD-Net            & 14.36 & 29.59 & 40.00 & 7.23 \\
IDE+PCB-Net        & 12.84 & 26.47 & 35.14 & 6.09 \\\hline
IDE+         & 7.75 & 20.09 & 27.43 & 3.38 \\
IDF$_{IDE+}$               & 13.63 & 30.05 & 38.95 & 6.63 \\\hline
AGW         & 12.50 & 24.40 & 30.90 & 6.10 \\
IDF$_{AGW}$           & 16.00 & 29.30 & 38.10 & 8.90\\ \hline
TransReID            & 16.00 & 31.10 & 39.90 & 8.40 \\
IDF$_{TransReID}$             & 17.20 & 34.40 & 43.80 & 9.20 \\
\hline
\end{tabular}
\end{table}

\subsection{Nighttime Re-ID Evaluation and Comparisons}
\noindent{\bf Evaluation on \emph{Night600}.} We evaluate six representative state-of-the-art methods on our \emph{Night600} dataset, including IDE+~\cite{IDE+_2019}, AGW~\cite{AGW_2021}, ABD-Net~\cite{ABDNet_2019}, TransReID~\cite{TransReID_2021}, IDE+PCB-Net~\cite{PCB_2018} and BoT~\cite{BoT_2019}.
From the results of Table~\ref{tab:sota_compare_exp}, it can be seen that the classic striping-based method IDE+PCB-Net, advanced transformer-based method TransReID and attention-based method ABD-Net achieve better performance in comparisons, which demonstrates that these methods can be more effective for the nighttime Re-ID compared with other methods in the comparison.
Since the backbone in MBranch and IEBranch can be replaced with an off-the-shelf person Re-ID method, our method can be regarded as a general framework.
Therefore, we select one classic Re-ID method IDE+, and two recently released Re-ID methods AGW and TransReID as the backbone embedded in our framework.

As shown in Table~\ref{tab:sota_compare_exp}, the classic IDE+ can achieve a comparable performance compared with ABD-Net when the proposed IDF framework is adopted (\textit{i.e.}, IDF$_{IDE+}$ \textit{vs.} ABD-Net).
In detail, our IDF$_{IDE+}$ outperforms the IDE+ by $5.88\%$ in rank-1 and $3.25$ in mAP, which proves the effectiveness of our approach to the real-world nighttime Re-ID.
Competitive nighttime Re-ID performance is also achieved by IDF$_{AGW}$ when compared with the state-of-the-art traditional person Re-ID method TransReID.
Furthermore, to validate the generalization of our framework, we also embed the backbone of transformer-based Re-ID algorithm (\emph{i.e.}, TransReID) to our framework, which can achieve the best performance on \emph{Night600}.
It is worth noting that since the training of Vision Transformer (ViT) based models relies on large-scale training dataset~\cite {bai2022improving}, while the scales of existing nighttime Re-ID datasets are not large to train TransReID well. 
Therefore, our method achieves a relatively small improvement (1.2\%) in rank-1 based on TransReID.
%
%
These experimental results show that our IDF can elevate the performance of traditional Re-ID methods when the task becomes nighttime person Re-ID.
\begin{table}[h]
\centering
\caption{Performance of three low-illumination enhancement methods with existing Re-ID methods on the \emph{Night600} dataset.}
\label{tab:enhance_compare_exp}
\begin{tabular}{cccccc}
\hline
                     & R-1  & R-5   & R-10  & $m$AP  \\\hline
IDE+             & $7.75$ & $20.09$ & $27.43$ & $3.38$ \\
w/ Ada Gamma  & $8.81$ & $20.05$ & $28.07$ & $3.59$ \\
w/ En-GAN   & $8.44$ & $20.69$ & $28.12$ & $3.54$ \\
w/ DCE   & $9.08$ & $21.61$ & $29.40$ & $3.71$ \\\hline
AGW             & $12.50$ & $24.40$ & $30.90$ & $6.10$ \\
w/ Ada Gamma             & $13.40$ & $24.20$ & $32.40$ & $6.10$ \\
w/ En-GAN              & $12.60$ & $24.80$ & $33.50$ & $5.40$ \\
w/ DCE              & $10.60$ & $22.40$ & $30.50$ & $4.70$ \\\hline
\hline
\end{tabular}
\end{table}

In addition, we also select three different enhancement methods: Ada Gamma~\cite{rahman2016adaptive}, En-GAN~\cite{ENGAN_2021} and DCE~\cite{DCE_2020}, to preprocess the nighttime person images and then feed them into two common daytime Re-ID algorithms ({\emph{i.e.}, IDE+ and AGW}) without jointly train the illumination enhancement network and the Re-ID model.
As shown in Table~\ref{tab:enhance_compare_exp}, the performance of IDE+ (w/ DCE) is increased compared with IDE+. 
Contradictorily, AGW (W/ DCE) can damage the performance of AGW. 
This result indicates that without jointly training the illumination enhancement network and the Re-ID model, the performance is unstable in nighttime person Re-ID.
Therefore, our IDF which jointly trains the illumination enhancement network and all the other components can get more stable improvement for nighttime person Re-ID. 
\begin{table}[]
\centering
\caption{Performance of representative Re-ID methods on the \emph{Knight} dataset.}
\label{tab:knigth_exp}
\begin{tabular}{llllll}
\hline
\multicolumn{1}{c}{\multirow{2}{*}{Setting}} & \multicolumn{1}{c}{\multirow{2}{*}{Methods}} & \multicolumn{4}{c}{Knight} \\ \cline{3-6} 
\multicolumn{1}{c}{}       & \multicolumn{1}{c}{} & R-1  & R-5  & R-10 & mAP  \\ \hline
\multirow{3}{*}{cam1-cam2} & Knight$_{Re-ID}$                & 11.7 & 16.7 & 23.2 & 8.3  \\
                           & IDE+                 & 14.6 & 22.8 & 27.3 & 10.0  \\
                           & IDF$_{IDE+}$       & 17.6 & 24.4 & 28.2 & 12.2 \\ \hline
\multirow{3}{*}{cam2-cam3} & Knight$_{Re-ID}$              & 14.3 & 22.5 & 26.7 & 10.2 \\
                           & IDE+                 & 19.2 & 25.7 & 29.3 & 13.5 \\
                           & IDF$_{IDE+}$        & 19.9 & 27.3 & 31.4 & 14.8 \\ \hline
\multirow{3}{*}{cam1-cam3} & Knight$_{Re-ID}$                & 6.3  & 16.5 & 23.4 & 5.5  \\
                           & IDE+                 & 5.8  & 10.6 & 13.6 & 5.8  \\
                           & IDF$_{IDE+}$        & 7.7  & 13.2 & 17.1 & 6.2  \\ \hline                
\end{tabular}
\end{table}

\noindent{\bf Evaluation on \emph{Knight}.} As mentioned in~\cite{Zhang2019NightPR}, we also find that the Knight dataset is very hard to train deep learning models, even with ResNet50, it usually takes one week to train a converged model on our device.
Considering the limitation of computing resources, we only evaluate the method proposed in~\cite{Zhang2019NightPR} (denoted as Knight$_{Re-ID}$), and our IDF method when IDE+ is used as the backbone of MBranch and IEBranch.
As reported in Table~\ref{tab:knigth_exp}, it shows that our approach can achieve the best performance on three settings in the \emph{Knight} dataset.
Moreover, the number of identities in the setting of cam2-cam3 is similar to our dataset, while IDE+ obtains lower performance on our \emph{Night600} dataset than the \emph{Knight} dataset.
This result demonstrates our dataset is more challenging compared with the \emph{Knight} dataset. 

\begin{table}[]
\centering
\caption{Ablation study on the \emph{Night600} dataset.}
\label{tab:ablation_exp}
\begin{tabular}{lllll}
\hline
\multicolumn{1}{c}{\multirow{2}{*}{Methods}} & \multicolumn{4}{c}{Night600} \\ \cline{2-5} 
\multicolumn{1}{c}{}                         & R-1   & R-5   & R-10  & mAP  \\ \hline
MB                                           & 7.75  & 20.09 & 27.43 & 3.38 \\
MB+MB (W/ cat)                                      & 9.26  & 21.56 & 29.86 & 3.80 \\ \hline
IEB                                          & 10.73 & 24.08 & 32.62 & 4.42 \\
MB+IEB (W/ cat)                                        & 10.00 & 21.93 & 31.19 & 4.47 \\ \hline
MB+IEB+IDM                                   & 13.03 & 27.75 & 36.79 & 6.48 \\
MB+IEB+IDM (W/ IFD)                              & 13.63 & 30.05 & 38.95 & 6.63 \\
\hline
\end{tabular}
\end{table}

\begin{table}[]
\centering
\caption{Experimental results of $IDF_{IDE+}$ on \emph{Night600} with different $\lambda_1$ and $\lambda_2$ values, where vertical and horizontal represent $\lambda_1$ and $\lambda_2$, respectively. }
\label{tab:Parameters-exp}
\begin{tabular}{l|lllll}
\hline
R-1   & 0.1        & 0.3        & 0.5        & 0.7        & 0.9        \\ \hline
0.1 & 13.63 & 12.80 & 13.21 & 11.74& 11.84 \\ \hline
0.3 & 11.88 & 11.01 & 12.57 & 11.38 & 12.39 \\ \hline
0.5 & 12.62 & 13.12 & 12.65 &  11.38  & 12.13 \\ \hline
0.7 & 11.29 & 11.42 & 12.56 & 11.92  & 11.78  \\ \hline
0.9 & 11.51 & 12.76 & 12.05 &  11.18 & 10.93  \\ \hline
\end{tabular}
\end{table}

\begin{figure}[h]
    \centering
    \includegraphics[scale=0.43]{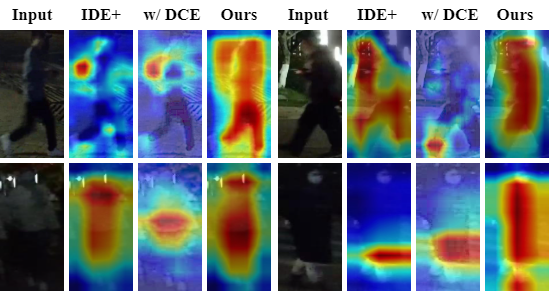} 
    \vspace{-.3cm}
    \caption{Visualizations on the activation map of ours and two baseline methods on \emph{Night600} dataset.}
    \label{fig:ours_cam}
\end{figure}
\begin{figure*}[h]
    \centering
    \includegraphics[scale=0.34]{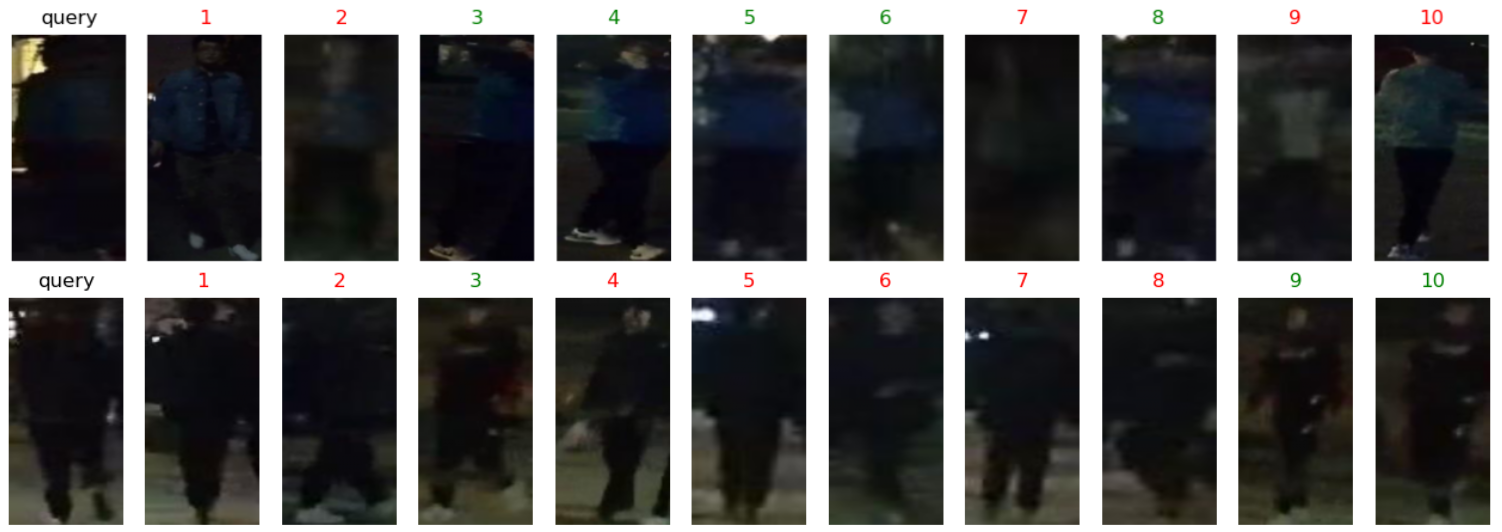}
    \vspace{-.3cm}
    \caption{ Sample retrieval results generated by the method of $IDF_{IDE+}$ on \emph{Night600}. Here the {\color{green} green} and {\color{red} red} numbers represent true positive and false positive samples, respectively.}
    \label{fig:rank_list}
\end{figure*}

\begin{figure}[h]
    \centering
    \includegraphics[scale=0.65]{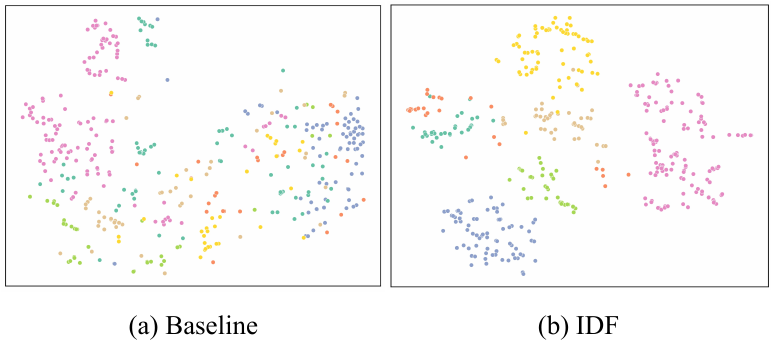}
    \vspace{-.3cm}
    \caption{ T-SNE~\cite{tSNE_hinton} visualization of IDF and baseline feature distribution. Different colors represent different IDs.}
    \label{fig:t_nse_idf}
\end{figure}

\subsection{Analysis}
\noindent{\bf Ablation Studies.} To verify the effectiveness of different components in our framework, we use ablation studies on MBranch, IEBranch, and IDModule with the proposed \emph{Night600} dataset.
Herein, we use MB, IEB, and IDM to represent the MBranch, IEBranch, and IDModule for short.
As shown in Table~\ref{tab:ablation_exp}, it can be seen that MB, which is a baseline method IDE+, obtains the lowest performance on the proposed dataset.
Considering the structure of IDF is two branches, we use two MBranch to build a two-stream network and employ concatenation operation to achieve fusion, denoted MB+MB (W/ cat), for a more fair comparison.
The result shows that simply doubling the number of parameters can only slightly improve the performance, which further proves the effectiveness of our MB+IEB (W/ cat). 
To prove the effectiveness of IEBranch, we also evaluate the performance of IEBranch on \emph{Night600}, denoted IEB, which achieves significant improvement over MB and MB+MB. 
In addition, we utilize a simple fusion method (\emph{i.e.}, concatenation) for fusing features of two branches (\emph{i.e.}, MBranch and IEBranch), denoted MB+IEB (W/ cat), to validate the effectiveness of IDM.
It leads to a worse result than using IEBranch only, which reveals that features of the two branches contain noise that can damage the performance.
Compared with the simple fusion method, our IDModule brings significant performance improvement (\emph{i.e.} MB+IEB+IDM), which demonstrates that our fusion strategy can be more effective for nighttime person Re-ID.
Finally, the illumination fused distillation strategy is introduced during training, denoted IFD, and the performance of our method (\emph{i.e.} MB+IEB+IDM (W/ IFD)) can be further improved (\emph{i.e.}, 0.6\% in Rank-1 and 2.3\% in rank-5).
%

\noindent{\bf Parameter Analysis.} As shown in Eq.~\ref{eq::12}, the hyper-parameters $\lambda_1$ and $\lambda_2$ are used to control influences of reconstruction loss and distillation loss.
To verify the robustness of our method, we try to change the values of $\lambda_1$ and $\lambda_2$.
As shown in Table~\ref{tab:Parameters-exp}, it can see that although using different $\lambda_1$ and $\lambda_2$ can affect the final performance to certain extent, it still shows better performance than baseline method on the \emph{Night600} dataset. 
In addition, with the change of $\lambda_1$ and $\lambda_2$, the performance of our method does not change a lot, which shows the robustness of our method.
Finally, we set two hyper-parameters to 0.1 and obtain the best performance from Table~\ref{tab:Parameters-exp}.
%

\noindent{\bf Visualizations.} To verify the effectiveness of the proposed IDF method, we draw activation maps of four persons from the \emph{Night600} dataset by utilizing GardCAM~\cite{Selvaraju2019GradCAMVE} on our method and two baseline methods.
Herein, the IDE+ algorithm is directly applied to the \emph{Night600} dataset as one baseline (\emph{i.e.} IDE+).
Moreover, we enhance the \emph{Night600} dataset with DCE-Net in the pre-processing stage, which is then evaluated by IDE+ and used as another baseline (\emph{i.e.} w/DCE).
%
%
Fig.~\ref{fig:ours_cam} indicates that both the baseline method and the method using low-illumination enhancement data are hard to capture discriminative regions on the body.
Compared with the two baseline methods, our framework (\emph{i.e.} Ours) shows better results, which suggests that our method can learn more discriminative representations of persons in real-world nighttime scenarios. 

To further show the advantages of our method, we employ the t-SNE~\cite{tSNE_hinton} technique to visualize the feature embeddings extracted by the baseline method and IDF. Specifically, we randomly select 7 identity data from the gallery set, and then extract the feature embeddings corresponding to these data using the baseline and IDF, respectively. Finally, we adopt the t-SNE technique to obtain the feature distribution as shown in Fig.~\ref{fig:t_nse_idf}. It can be seen that IDF achieves better classification results for different identity data. As a result, our method shows a better identification advantage over the baseline method in nighttime Re-ID.

\begin{table}[]
\centering
\caption{Experimental results of IDF$_{IDE+}$ on the cross-domain settings of \emph{Knight} and \emph{Night600}.}
\label{tab:knight-night600-exp}
\begin{tabular}{@{}cccccc@{}}
\toprule
Train\_data               & Test\_data                & R-1  & R-5  & R-10 & mAP  \\ \midrule
Knight(c1c3)              & \multirow{3}{*}{Night600} & 0.23 & 0.60 & 0.60 & 0.89 \\
Knight(c1c2)              &                           & 1.38 & 4.45 & 6.74 & 0.53 \\
Knight(c2c3)              &                           & 1.42 & 4.00 & 7.02 & 0.55 \\ \midrule
\multirow{3}{*}{Night600} & Knight(c1c3)              & 1.95 & 4.89 & 7.34 & 2.56 \\
                          & Knight(c1c2)              & 2.12 & 4.06 & 5.74 & 1.38 \\
                          & Knight(c2c3)              & 2.49 & 5.61 & 8.09 & 1.50 \\ \bottomrule
\end{tabular}
\end{table}

\noindent{\bf Limitations of the method.} Although our method achieves better performance on two dataset comparisons and visualization experiments, it remains three major limitations.
First, since the two-branch structure inevitably introduces a large number of parameters, it will increase the computational burden and thus downgrade the practicality in real applications. 
To handle this issue, we will explore a more efficient single-branch network in the future, for example, study an input-aware feature extractor to adaptively model person representations under various illumination conditions at night.
Second, our framework is hard to handle the negative samples with a similar appearance to the query image, as shown in Fig~\ref{fig:rank_list}.
To this end, we will subsequently investigate a more effective illumination enhancement scheme to highlight the discriminative  features of the person images to better distinguish similar samples.
Third, we perform cross-domain experiments on two real nightly Re-ID datasets. As shown in Table~\ref{tab:knight-night600-exp}, it can be seen that the proposed approach is still hard to handle the challenge of large inter-domain differences. 
In the future, we will consider introducing domain generalization techniques to improve performance in this setting.

\section{Conclusion}

We propose a novel illumination distillation framework (IDF) to promote representation learning for nighttime person Re-ID.
The proposed method can effectively leverage discriminative information from both nighttime images and their corresponding illumination-enhanced images for real-world nighttime person Re-ID.
Meanwhile, we contribute a new dataset called \emph{Night600} for the nighttime person Re-ID community.
Through comprehensive experiments, we show the effectiveness of the proposed IDF.
In future work, we consider domain adaptation technology to improve the performance of nighttime person Re-ID by leveraging daytime person Re-ID data. 
In addition, inspired by~\cite{shen2021distilled,zhao2021real}, we will also develop more efficient lightweight weight networks while maintaining performance without degradation.

\bibliographystyle{IEEEtran}
\bibliography{TMM}

\begin{thebibliography}{10}
\providecommand{\url}[1]{#1}
\csname url@samestyle\endcsname
\providecommand{\newblock}{\relax}
\providecommand{\bibinfo}[2]{#2}
\providecommand{\BIBentrySTDinterwordspacing}{\spaceskip=0pt\relax}
\providecommand{\BIBentryALTinterwordstretchfactor}{4}
\providecommand{\BIBentryALTinterwordspacing}{\spaceskip=\fontdimen2\font plus
\BIBentryALTinterwordstretchfactor\fontdimen3\font minus
  \fontdimen4\font\relax}
\providecommand{\BIBforeignlanguage}[2]{{%
\expandafter\ifx\csname l@#1\endcsname\relax
\typeout{** WARNING: IEEEtran.bst: No hyphenation pattern has been}%
\typeout{** loaded for the language `#1'. Using the pattern for}%
\typeout{** the default language instead.}%
\else
\language=\csname l@#1\endcsname
\fi
#2}}
\providecommand{\BIBdecl}{\relax}
\BIBdecl

\bibitem{night_report}
T.~E. Team, ``Crimes that happen while you sleep,''
  \emph{https://www.thesleepjudge.com/crimes-that-happen-while-you-sleep/},
  2021.

\bibitem{gao2020dcr}
Z.~Gao, L.~Gao, H.~Zhang, Z.~Cheng, R.~Hong, and S.~Chen, ``Dcr: A unified
  framework for holistic/partial person reid,'' \emph{IEEE Transactions on
  Multimedia}, vol.~23, pp. 3332--3345, 2020.

\bibitem{CPA_tifs}
D.~Wu, M.~Ye, G.~Lin, X.~Gao, and J.~Shen, ``Person re-identification by
  context-aware part attention and multi-head collaborative learning,''
  \emph{IEEE Transactions on Information Forensics and Security}, vol.~17, pp.
  115--126, 2022.

\bibitem{AIIS_pami}
M.~Ye, J.~Shen, X.~Zhang, P.~C. Yuen, and S.-F. Chang, ``Augmentation invariant
  and instance spreading feature for softmax embedding,'' \emph{IEEE
  Transactions on Pattern Analysis and Machine Intelligence}, vol.~44, no.~2,
  pp. 924--939, 2022.

\bibitem{CR_tip}
M.~Ye, H.~Li, B.~Du, J.~Shen, L.~Shao, and S.~C.~H. Hoi, ``Collaborative
  refining for person re-identification with label noise,'' \emph{IEEE
  Transactions on Image Processing}, vol.~31, pp. 379--391, 2022.

\bibitem{adaptive_ill_2019TMM}
Z.~Zeng, Z.~Wang, Z.~Wang, Y.-Y. Chuang, and S.~Satoh, ``Illumination-adaptive
  person re-identification,'' \emph{IEEE TRANSACTIONS ON MULTIMEDIA}, vol.~22,
  no.~12, pp. 3064--3074, 2020.

\bibitem{BoT_2019}
H.~Luo, W.~Jiang, Y.~Gu, F.~Liu, X.~Liao, S.~Lai, and J.~Gu, ``A strong
  baseline and batch normalization neck for deep person re-identification,''
  \emph{IEEE Transactions on Multimedia}, vol.~22, no.~10, pp. 2597--2609,
  2019.

\bibitem{jia2022learning}
M.~Jia, X.~Cheng, S.~Lu, and J.~Zhang, ``Learning disentangled representation
  implicitly via transformer for occluded person re-identification,''
  \emph{IEEE Transactions on Multimedia}, 2022.

\bibitem{zhang2020person}
S.~Zhang, Q.~Zhang, Y.~Yang, X.~Wei, P.~Wang, B.~Jiao, and Y.~Zhang, ``Person
  re-identification in aerial imagery,'' \emph{IEEE Transactions on
  Multimedia}, vol.~23, pp. 281--291, 2020.

\bibitem{ABDNet_2019}
T.~Chen, S.~Ding, J.~Xie, Y.~Yuan, W.~Chen, Y.~Yang, Z.~Ren, and Z.~Wang,
  ``Abd-net: Attentive but diverse person re-identification,'' in
  \emph{Proceedings of the IEEE International Conference on Computer Vision},
  2019, pp. 8351--8361.

\bibitem{AGW_2021}
M.~Ye, J.~Shen, G.~Lin, T.~Xiang, L.~Shao, and S.~C.~H. Hoi, ``Deep learning
  for person re-identification: A survey and outlook,'' \emph{IEEE Transactions
  on Pattern Analysis and Machine Intelligence}, pp. 1--1, 2021.

\bibitem{TransReID_2021}
S.~He, H.~Luo, P.~Wang, F.~Wang, H.~Li, and W.~Jiang, ``Transreid:
  Transformer-based object re-identification,'' in \emph{Proceedings of the
  IEEE International Conference on Computer Vision}, 2021, pp.
  14\,993--15\,002.

\bibitem{PCB_2018}
Y.~Sun, L.~Zheng, Y.~Yang, Q.~Tian, and S.~Wang, ``Beyond part models: Person
  retrieval with refined part pooling (and a strong convolutional baseline),''
  in \emph{Proceedings of the European Conference on Computer Vision}, 2018.

\bibitem{IDE+_2019}
Z.~Zheng, L.~Zheng, and Y.~Yang, ``A discriminatively learned cnn embedding for
  person reidentification,'' \emph{Transactions on Multimedia Computing,
  Communications and Applications}, vol.~14, pp. 1 -- 20, 2018.

\bibitem{Real_world_CVPR2020}
Y.~Huang, Z.~Zha, X.~Fu, R.~Hong, and L.~Li, ``Real-world person
  re-identification via degradation invariance learning,'' in \emph{Proceedings
  of the IEEE Conference on Computer Vision and Pattern Recognition}, 2020, pp.
  14\,072--14\,082.

\bibitem{Ill_Invariant_2019ACMMM}
Y.~Huang, Z.~Zha, X.~Fu, and W.~Zhang, ``Illumination-invariant person
  re-identification,'' in \emph{Proceedings of the ACM International Conference
  on Multimedia}, 2019, p. 365–373.

\bibitem{IllAP_2020icip}
Z.~Zhang, R.~Y.~D. Xu, S.~Jiang, Y.~Li, C.~Huang, and C.~Deng, ``Illumination
  adaptive person reid based on teacher-student model and adversarial
  training,'' in \emph{Proceedings of the International Conference on Image
  Processing}, 2020, pp. 2321--2325.

\bibitem{market1501_2015}
L.~Zheng, L.~Shen, L.~Tian, S.~Wang, J.~Wang, and Q.~Tian, ``Scalable person
  re-identification: A benchmark,'' in \emph{Proceedings of the IEEE
  International Conference on Computer Vision}, 2015, pp. 1116--1124.

\bibitem{zhang2019kindling}
Y.~Zhang, J.~Zhang, and X.~Guo, ``Kindling the darkness: A practical low-light
  image enhancer,'' in \emph{Proceedings of the 27th ACM international
  conference on multimedia}, 2019, pp. 1632--1640.

\bibitem{lu2020tbefn}
K.~Lu and L.~Zhang, ``Tbefn: A two-branch exposure-fusion network for low-light
  image enhancement,'' \emph{IEEE Transactions on Multimedia}, vol.~23, pp.
  4093--4105, 2020.

\bibitem{yang2020fidelity}
W.~Yang, S.~Wang, Y.~Fang, Y.~Wang, and J.~Liu, ``From fidelity to perceptual
  quality: A semi-supervised approach for low-light image enhancement,'' in
  \emph{Proceedings of the IEEE/CVF conference on computer vision and pattern
  recognition}, 2020, pp. 3063--3072.

\bibitem{triantafyllidou2020low}
D.~Triantafyllidou, S.~Moran, S.~McDonagh, S.~Parisot, and G.~Slabaugh, ``Low
  light video enhancement using synthetic data produced with an intermediate
  domain mapping,'' in \emph{Proceedings of the European Conference on Computer
  Vision}, 2020, pp. 103--119.

\bibitem{wang2020lightening}
L.-W. Wang, Z.-S. Liu, W.-C. Siu, and D.~P. Lun, ``Lightening network for
  low-light image enhancement,'' \emph{IEEE Transactions on Image Processing},
  vol.~29, pp. 7984--7996, 2020.

\bibitem{DCE_2020}
C.~Guo, C.~Li, J.~Guo, C.~C. Loy, J.~Hou, S.~Kwong, and R.~Cong,
  ``Zero-reference deep curve estimation for low-light image enhancement,'' in
  \emph{Proceedings of the IEEE Conference on Computer Vision and Pattern
  Recognition}, 2020, pp. 1780--1789.

\bibitem{dce++tpami}
C.~Li, C.~Guo, and C.~C. Loy, ``Learning to enhance low-light image via
  zero-reference deep curve estimation,'' \emph{IEEE Transactions on Pattern
  Analysis and Machine Intelligence}, vol.~44, no.~8, pp. 4225--4238, 2021.

\bibitem{he2016deep}
K.~He, X.~Zhang, S.~Ren, and J.~Sun, ``Deep residual learning for image
  recognition,'' in \emph{Proceedings of the IEEE Conference on Computer Vision
  and Pattern Recognition}, 2016, pp. 770--778.

\bibitem{ENGAN_2021}
Y.~Jiang, X.~Gong, D.~Liu, Y.~Cheng, C.~Fang, X.~Shen, J.~Yang, P.~Zhou, and
  Z.~Wang, ``Enlightengan: Deep light enhancement without paired supervision,''
  \emph{IEEE Transactions on Image Processing}, vol.~30, pp. 2340--2349, 2021.

\bibitem{CMMCCA_iccv}
X.~Hao, S.~Zhao, M.~Ye, and J.~Shen, ``Cross-modality person re-identification
  via modality confusion and center aggregation,'' in \emph{2021 IEEE/CVF
  International Conference on Computer Vision (ICCV)}, 2021, pp.
  16\,383--16\,392.

\bibitem{vireid_trimodal_tifs}
M.~Ye, J.~Shen, and L.~Shao, ``Visible-infrared person re-identification via
  homogeneous augmented tri-modal learning,'' \emph{IEEE Transactions on
  Information Forensics and Security}, vol.~16, pp. 728--739, 2021.

\bibitem{DTRM_tifs}
M.~Ye, C.~Chen, J.~Shen, and L.~Shao, ``Dynamic tri-level relation mining with
  attentive graph for visible infrared re-identification,'' \emph{IEEE
  Transactions on Information Forensics and Security}, vol.~17, pp. 386--398,
  2022.

\bibitem{Zhang2019NightPR}
J.~Zhang, Y.~Yuan, and Q.~Wang, ``Night person re-identification and a
  benchmark,'' \emph{IEEE Access}, vol.~7, pp. 95\,496--95\,504, 2019.

\bibitem{deng2018image}
W.~Deng, L.~Zheng, Q.~Ye, G.~Kang, Y.~Yang, and J.~Jiao, ``Image-image domain
  adaptation with preserved self-similarity and domain-dissimilarity for person
  re-identification,'' in \emph{Proceedings of the IEEE Conference on Computer
  Vision and Pattern Recognition}, 2018, pp. 994--1003.

\bibitem{suh2018part}
Y.~Suh, J.~Wang, S.~Tang, T.~Mei, and K.~M. Lee, ``Part-aligned bilinear
  representations for person re-identification,'' in \emph{Proceedings of the
  European Conference on Computer Vision}, 2018, pp. 402--419.

\bibitem{li2018harmonious}
W.~Li, X.~Zhu, and S.~Gong, ``Harmonious attention network for person
  re-identification,'' in \emph{Proceedings of the IEEE Conference on Computer
  Vision and Pattern Recognition}, 2018, pp. 2285--2294.

\bibitem{song2018mask}
C.~Song, Y.~Huang, W.~Ouyang, and L.~Wang, ``Mask-guided contrastive attention
  model for person re-identification,'' in \emph{Proceedings of the IEEE
  Conference on Computer Vision and Pattern Recognition}, 2018, pp. 1179--1188.

\bibitem{HAT_2021}
G.~Zhang, P.~Zhang, J.~Qi, and H.~Lu, ``Hat: Hierarchical aggregation
  transformers for person re-identification,'' in \emph{Proceedings of the ACM
  International Conference on Multimedia}, 2021, pp. 516--–525.

\bibitem{RAGA_2020}
Z.~Zhang, C.~Lan, W.~Zeng, X.~Jin, and Z.~Chen, ``Relation-aware global
  attention for person re-identification,'' in \emph{Proceedings of the IEEE
  Conference on Computer Vision and Pattern Recognition}, 2020, pp. 3183--3192.

\bibitem{wang2018resource}
Y.~Wang, L.~Wang, Y.~You, X.~Zou, V.~Chen, S.~Li, G.~Huang, B.~Hariharan, and
  K.~Q. Weinberger, ``Resource aware person re-identification across multiple
  resolutions,'' in \emph{Proceedings of the IEEE Conference on Computer Vision
  and Pattern Recognition}, 2018, pp. 8042--8051.

\bibitem{xu2020black}
B.~Xu, L.~He, X.~Liao, W.~Liu, Z.~Sun, and T.~Mei, ``Black re-id: A
  head-shoulder descriptor for the challenging problem of person
  re-identification,'' in \emph{Proceedings of the ACM International Conference
  on Multimedia}, 2020, pp. 673--681.

\bibitem{gao2020pose}
S.~Gao, J.~Wang, H.~Lu, and Z.~Liu, ``Pose-guided visible part matching for
  occluded person reid,'' in \emph{Proceedings of the IEEE Conference on
  Computer Vision and Pattern Recognition}, 2020, pp. 11\,744--11\,752.

\bibitem{huang2019celebrities}
Y.~Huang, Q.~Wu, J.~Xu, and Y.~Zhong, ``Celebrities-reid: A benchmark for
  clothes variation in long-term person re-identification,'' in
  \emph{IJCNN}.\hskip 1em plus 0.5em minus 0.4em\relax IEEE, 2019, pp. 1--8.

\bibitem{rahman2016adaptive}
S.~Rahman, M.~M. Rahman, M.~Abdullah-Al-Wadud, G.~D. Al-Quaderi, and
  M.~Shoyaib, ``An adaptive gamma correction for image enhancement,''
  \emph{EURASIP Journal on Image and Video Processing}, vol. 2016, no.~1, pp.
  1--13, 2016.

\bibitem{VIPeR}
D.~Gray, S.~Brennan, and H.~Tao, ``Evaluating appearance models for
  recognition, reacquisition, and tracking,'' in \emph{Proceedings of the IEEE
  international workshop on performance evaluation for tracking and
  surveillance}, 2007, pp. 1--7.

\bibitem{CUHK03}
W.~Li, R.~Zhao, T.~Xiao, and X.~Wang, ``Deepreid: Deep filter pairing neural
  network for person re-identification,'' \emph{Proceedings of the IEEE
  Conference on Computer Vision and Pattern Recognition}, pp. 152--159, 2014.

\bibitem{DukeMTMC}
Z.~Zheng, L.~Zheng, and Y.~Yang, ``Unlabeled samples generated by gan improve
  the person re-identification baseline in vitro,'' \emph{Proceedings of the
  IEEE International Conference on Computer Vision}, pp. 3774--3782, 2017.

\bibitem{Airport}
S.~Karanam, M.~Gou, Z.~Wu, A.~Rates-Borras, O.~I. Camps, and R.~J. Radke, ``A
  systematic evaluation and benchmark for person re-identification: Features,
  metrics, and datasets,'' \emph{IEEE Transactions on Pattern Analysis and
  Machine Intelligence}, vol.~41, pp. 523--536, 2019.

\bibitem{MSMT17}
L.~Wei, S.~Zhang, W.~Gao, and Q.~Tian, ``Person transfer gan to bridge domain
  gap for person re-identification,'' \emph{Proceedings of the IEEE Conference
  on Computer Vision and Pattern Recognition}, pp. 79--88, 2018.

\bibitem{RPIfield}
M.~Zheng, S.~Karanam, and R.~J. Radke, ``Rpifield: A new dataset for temporally
  evaluating person re-identification,'' \emph{Proceedings of the IEEE
  Conference on Computer Vision and Pattern Recognition Workshop}, pp.
  1974--19\,742, 2018.

\bibitem{hinton2015distilling}
G.~Hinton, O.~Vinyals, and J.~Dean, ``Distilling the knowledge in a neural
  network,'' \emph{arXiv preprint arXiv:1503.02531}, 2015.

\bibitem{huang2017like}
Z.~Huang and N.~Wang, ``Like what you like: Knowledge distill via neuron
  selectivity transfer,'' \emph{arXiv preprint arXiv:1707.01219}, 2017.

\bibitem{li2020few}
T.~Li, J.~Li, Z.~Liu, and C.~Zhang, ``Few sample knowledge distillation for
  efficient network compression,'' in \emph{Proceedings of the IEEE/CVF
  Conference on Computer Vision and Pattern Recognition}, 2020, pp.
  14\,639--14\,647.

\bibitem{heo2019knowledge}
B.~Heo, M.~Lee, S.~Yun, and J.~Y. Choi, ``Knowledge distillation with
  adversarial samples supporting decision boundary,'' in \emph{Proceedings of
  the AAAI conference on artificial intelligence}, vol.~33, no.~01, 2019, pp.
  3771--3778.

\bibitem{guo2020online}
Q.~Guo, X.~Wang, Y.~Wu, Z.~Yu, D.~Liang, X.~Hu, and P.~Luo, ``Online knowledge
  distillation via collaborative learning,'' in \emph{Proceedings of the
  IEEE/CVF Conference on Computer Vision and Pattern Recognition}, 2020, pp.
  11\,020--11\,029.

\bibitem{chung2020feature}
I.~Chung, S.~Park, J.~Kim, and N.~Kwak, ``Feature-map-level online adversarial
  knowledge distillation,'' in \emph{International Conference on Machine
  Learning}, 2020, pp. 2006--2015.

\bibitem{shen2021distilled}
J.~Shen, Y.~Liu, X.~Dong, X.~Lu, F.~S. Khan, and S.~Hoi, ``Distilled siamese
  networks for visual tracking,'' \emph{IEEE Transactions on Pattern Analysis
  and Machine Intelligence}, vol.~44, no.~12, pp. 8896--8909, 2021.

\bibitem{chollet2017xception}
F.~Chollet, ``Xception: Deep learning with depthwise separable convolutions,''
  in \emph{Proceedings of the IEEE conference on computer vision and pattern
  recognition}, 2017, pp. 1251--1258.

\bibitem{KD_tits_2}
S.~Hwang, J.~Lee, W.~J. Kim, S.~Woo, K.~Lee, and S.~Lee, ``Lidar depth
  completion using color-embedded information via knowledge distillation,''
  \emph{IEEE Transactions on Intelligent Transportation Systems}, pp. 1--15,
  2021.

\bibitem{onlinekl_2021iccv}
Z.~Li, J.~Ye, M.~Song, Y.~Huang, and Z.~Pan, ``Online knowledge distillation
  for efficient pose estimation,'' in \emph{Proceedings of the IEEE
  International Conference on Computer Vision}, 2021, pp. 11\,740--11\,750.

\bibitem{KD_tits_1}
S.~An, Q.~Liao, Z.~Lu, and J.-H. Xue, ``Efficient semantic segmentation via
  self-attention and self-distillation,'' \emph{IEEE Transactions on
  Intelligent Transportation Systems}, pp. 1--11, 2022.

\bibitem{wu2019detectron2}
Y.~Wu, A.~Kirillov, F.~Massa, W.-Y. Lo, and R.~Girshick, ``Detectron2,''
  \url{https://github.com/facebookresearch/detectron2}, 2019.

\bibitem{jiang2022unsupervised}
Q.~Jiang, Y.~Mao, R.~Cong, W.~Ren, C.~Huang, and F.~Shao, ``Unsupervised
  decomposition and correction network for low-light image enhancement,''
  \emph{IEEE Transactions on Intelligent Transportation Systems}, 2022.

\bibitem{hassaballah2020vehicle}
M.~Hassaballah, M.~A. Kenk, K.~Muhammad, and S.~Minaee, ``Vehicle detection and
  tracking in adverse weather using a deep learning framework,'' \emph{IEEE
  transactions on intelligent transportation systems}, vol.~22, no.~7, pp.
  4230--4242, 2020.

\bibitem{wang2022sfnet}
H.~Wang, Y.~Chen, Y.~Cai, L.~Chen, Y.~Li, M.~A. Sotelo, and Z.~Li, ``Sfnet-n:
  An improved sfnet algorithm for semantic segmentation of low-light autonomous
  driving road scenes,'' \emph{IEEE Transactions on Intelligent Transportation
  Systems}, 2022.

\bibitem{RGBNT201_aaai}
A.~Zheng, Z.~Wang, Z.~Chen, C.~Li, and J.~Tang, ``Robust multi-modality person
  re-identification,'' in \emph{Proceedings of the AAAI Conference on
  Artificial Intelligence}, vol.~35, no.~4, 2021, pp. 3529--3537.

\bibitem{deng2009imagenet}
J.~Deng, W.~Dong, R.~Socher, L.-J. Li, K.~Li, and L.~Fei-Fei, ``Imagenet: A
  large-scale hierarchical image database,'' in \emph{Proceedings of the IEEE
  Conference on Computer Vision and Pattern Recognition}, 2009, pp. 248--255.

\bibitem{ViT_2021}
A.~Dosovitskiy, L.~Beyer, A.~Kolesnikov, D.~Weissenborn, X.~Zhai,
  T.~Unterthiner, M.~Dehghani, M.~Minderer, G.~Heigold, S.~Gelly, J.~Uszkoreit,
  and N.~Houlsby, ``An image is worth 16x16 words: Transformers for image
  recognition at scale,'' in \emph{Proceedings of the International Conference
  on Learning Representations}, 2021.

\bibitem{bai2022improving}
J.~Bai, L.~Yuan, S.-T. Xia, S.~Yan, Z.~Li, and W.~Liu, ``Improving vision
  transformers by revisiting high-frequency components,'' 2022.

\bibitem{tSNE_hinton}
L.~Van~der Maaten and G.~Hinton, ``Visualizing data using t-sne.''
  \emph{Journal of machine learning research}, vol.~9, no.~11, 2008.

\bibitem{Selvaraju2019GradCAMVE}
R.~R. Selvaraju, A.~Das, R.~Vedantam, M.~Cogswell, D.~Parikh, and D.~Batra,
  ``Grad-cam: Visual explanations from deep networks via gradient-based
  localization,'' \emph{International Journal of Computer Vision}, vol. 128,
  pp. 336--359, 2019.

\bibitem{zhao2021real}
Z.~Zhao, S.~Zhao, and J.~Shen, ``Real-time and light-weighted unsupervised
  video object segmentation network,'' \emph{Pattern Recognition}, vol. 120, p.
  108120, 2021.

\end{thebibliography}


 




\vfill

\end{document}